\documentclass[11pt]{article}

\usepackage[preprint]{acl}
\usepackage{booktabs}
\usepackage{multirow}
\usepackage{amsmath}
\usepackage{times}
\usepackage{latexsym}
\usepackage[table]{xcolor}
\usepackage{tabularx}

\usepackage{pgfplots}
\pgfplotsset{compat=1.18}
\usepackage{subcaption}

\usepackage{algorithm}
\usepackage{algorithmic}
\usepackage{float}

\usepackage{graphicx}
\usepackage{tikz}
\usepackage{pgfplots}
\pgfplotsset{compat=1.18}

\usetikzlibrary{plotmarks}
\usetikzlibrary{pgfplots.groupplots}
\usetikzlibrary{calc}
\usetikzlibrary{positioning}

\newcommand{\googlelogo}{Google.jpg}
\newcommand{\metalogo}{meta.png}

\newcommand{\mistrallogo}{mistral.png}
\newcommand{\allenailogo}{allenai.png}
\newcommand{\nvidialogo}{nvidia.png}

\newcommand{\qwenlogo}{qwen-icon.png}
\newcommand{\tiilogo}{tii.jpg}
\newcommand{\essentiallogo}{essentialai.png}
\newcommand{\milogo}{xiaomi.png}
\newcommand{\qalblogo}{robot.png}

\usepackage[T1]{fontenc}

\usepackage[utf8]{inputenc}

\usepackage{microtype}

\usepackage{inconsolata}

\usepackage{graphicx}

\usepackage{algorithmic}

\title{UrduBench: An Urdu Reasoning Benchmark using Contextually Ensembled Translations with Human-in-the-Loop}

\author{
\textbf{Muhammad Ali Shafique\textsuperscript{1}},
 \textbf{Areej Mehboob\textsuperscript{1}},
 \textbf{Layba Fiaz\textsuperscript{1}},\\
 \textbf{Muhammad Usman Qadeer\textsuperscript{1}},
  \textbf{Hamza Farooq\textsuperscript{1}}
\\
 \textsuperscript{1}Traversaal.ai,
 \small{
   \textbf{Correspondence:} \href{mailto:email@domain}{ali@traversaal.ai}
 }
}

\begin{document}
\maketitle
\begin{abstract}

Recent advances in large language models (LLMs) have led to strong reasoning capabilities; however, evaluating such models in low-resource languages remains challenging due to the lack of standardized benchmarks. In particular, Urdu reasoning evaluation has been limited by the sensitivity of machine translation and an emphasis on general language tasks rather than reasoning benchmarks. In this paper, we propose a contextually ensembled translation framework with human-in-the-loop validation that leverages multiple translation systems to develop Urdu reasoning benchmarks while preserving contextual and structural integrity. Using this framework, we translate widely adopted reasoning and question-answering benchmarks, including MGSM, MATH-500, CommonSenseQA, and OpenBookQA, into Urdu, collectively referred to as UrduBench, and conduct a comprehensive evaluation of both reasoning-oriented and instruction-tuned LLMs across multiple prompting strategies. Our analysis reveals performance differences across (1) four datasets, (2) five task difficulty levels, (3) diverse model architectures, (4) multiple model scaling settings, and (5) language consistency tests. We find that multi-step and symbolic reasoning tasks pose significant challenges in Urdu, and that stable language alignment is a critical prerequisite for robust reasoning. Overall, our work establishes a scalable methodology for standardized reasoning evaluation in Urdu and provides empirical insights into multilingual reasoning failures. This experimental setup is also broadly applicable to other low-resource languages. The code and datasets will be publicly released.

\end{abstract}

\section{Introduction}

\begin{table*}[h!]
    \centering
    \includegraphics[width=0.95\textwidth]{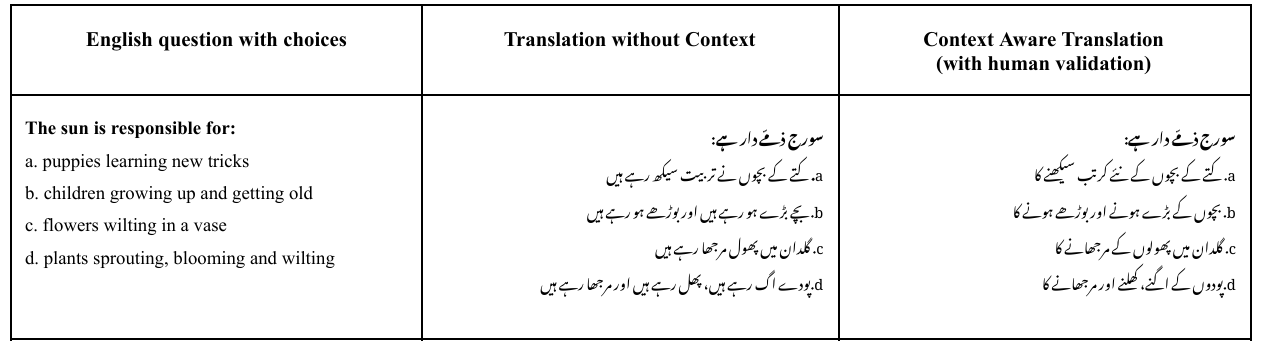}
    \caption{Example with and without Context Translation}
    \label{tab:one_example_context}
\end{table*}











Large Language Models (LLMs) have demonstrated remarkable performance across reasoning, comprehension, and knowledge-intensive tasks. However, the benefits of these advancements are not evenly distributed across languages. Low-resource languages such as Urdu continue to face a significant gap in standardized evaluation resources. This disparity not only limits the development of Urdu-capable LLMs but also restricts fair and rigorous comparison across multilingual models.

\begin{figure}[t!]
\centering
\begin{tikzpicture}

\begin{axis}[
    width=0.5\textwidth,
    height=0.4\textwidth,
    xlabel={Model Parameters (Billions)},
    ylabel={Average Accuracy (\%)},
    xmin=0, xmax=16,
    xtick={0, 2, 4, 6, 8, 10, 12, 14, 16},
    ymin=15, ymax=62,
    ytick={20, 30, 40, 50, 60},
    grid=major,
    grid style={dashed, gray!30},
    legend style={
        at={(0.98,0.02)},
        anchor=south east,
        fill=white,
        fill opacity=0.9,
        draw=black!50,
        rounded corners=2pt,
        font=\small,
        cells={anchor=west},
        line width=0.8pt,
    },
    tick label style={font=\footnotesize},
    label style={font=\normalsize},
    every axis plot/.append style={thick},
    xlabel style={font=\footnotesize},
    ylabel style={font=\footnotesize},
]

\def\logosize{0.28cm}


\node[anchor=center] at (axis cs:4, 44.9) {\includegraphics[width=0.20cm]{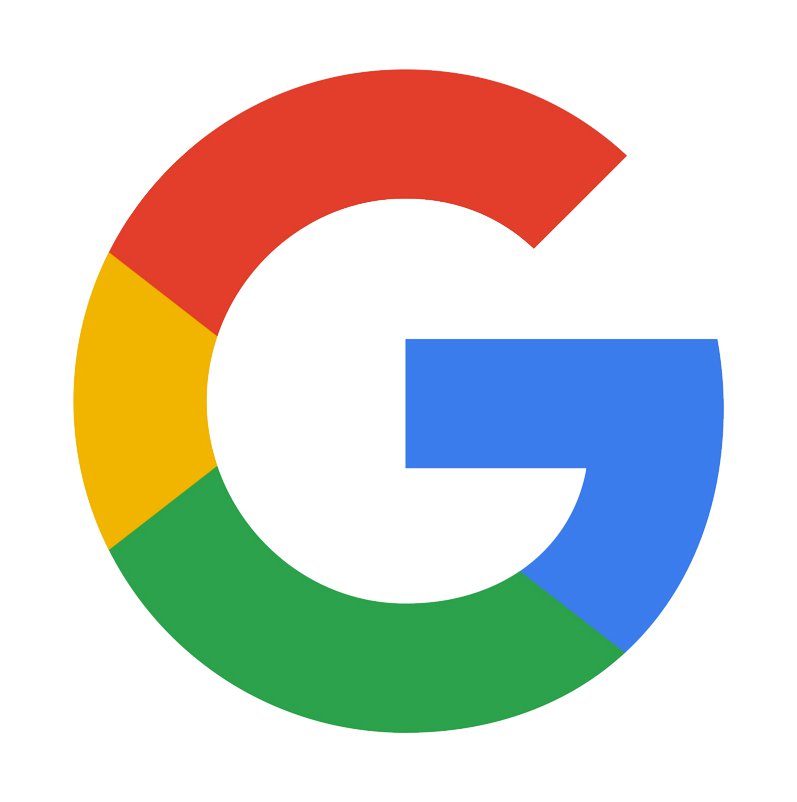}};
\node[anchor=center] at (axis cs:9, 42.2) {\includegraphics[width=0.20cm]{\googlelogo}};
\node[anchor=center] at (axis cs:12, 59.4) {\includegraphics[width=0.20cm]{\googlelogo}};

\node[anchor=center] at (axis cs:1, 17.0) {\includegraphics[width=0.20cm]{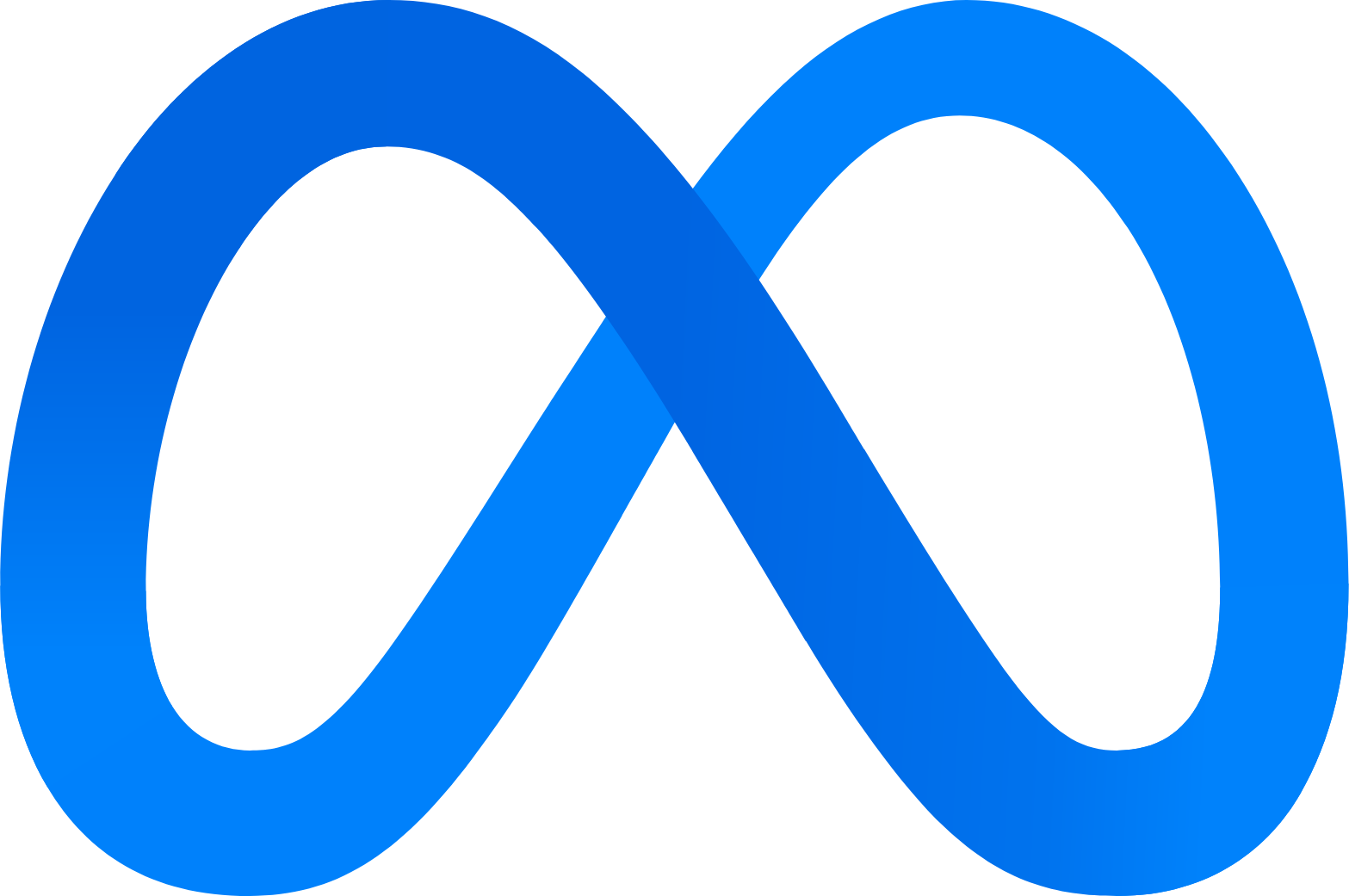}};
\node[anchor=center] at (axis cs:3, 29.7) {\includegraphics[width=0.20cm]{\metalogo}};
\node[anchor=center] at (axis cs:8, 42.9) {\includegraphics[width=0.20cm]{\metalogo}};

\node[anchor=center] at (axis cs:1.5, 19.6) {\includegraphics[width=0.20cm]{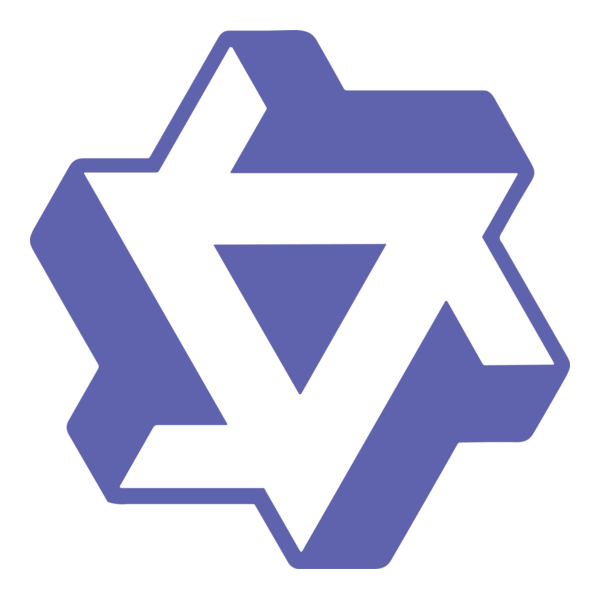}};
\node[anchor=center] at (axis cs:7, 38.3) {\includegraphics[width=0.20cm]{\qwenlogo}};
\node[anchor=center] at (axis cs:14, 44.9) {\includegraphics[width=0.20cm]{\qwenlogo}};

\node[anchor=center] at (axis cs:7, 17.2) {\includegraphics[width=0.20cm]{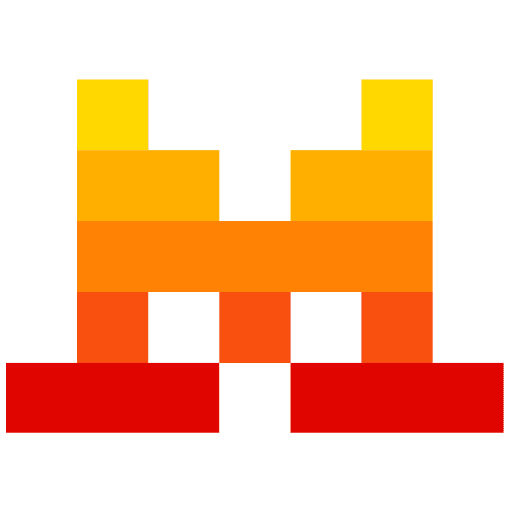}};
\node[anchor=center] at (axis cs:8, 21.4) {\includegraphics[width=0.20cm]{\mistrallogo}};
\node[anchor=center] at (axis cs:8, 34.6) {\includegraphics[width=0.20cm]{\mistrallogo}};

\node[anchor=center] at (axis cs:7.1, 23.7) {\includegraphics[width=0.35cm]{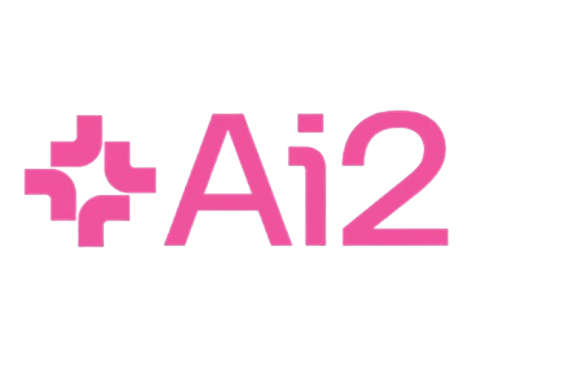}};
\node[anchor=center] at (axis cs:8, 27.6) {\includegraphics[width=\logosize]{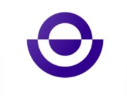}};
\node[anchor=center] at (axis cs:4, 24.5) {\includegraphics[width=\logosize]{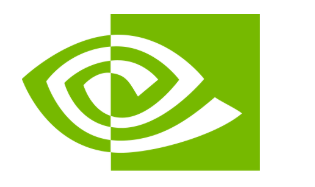}};
\node[anchor=center] at (axis cs:7, 53.5) {\includegraphics[width=\logosize]{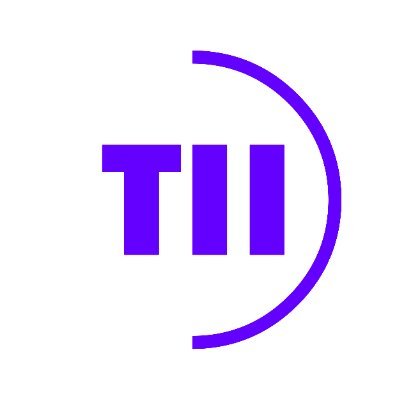}};
\node[anchor=center] at (axis cs:7, 40.2) {\includegraphics[width=0.20cm]{\qwenlogo}};
\node[anchor=center] at (axis cs:7, 25.6) {\includegraphics[width=0.20cm]{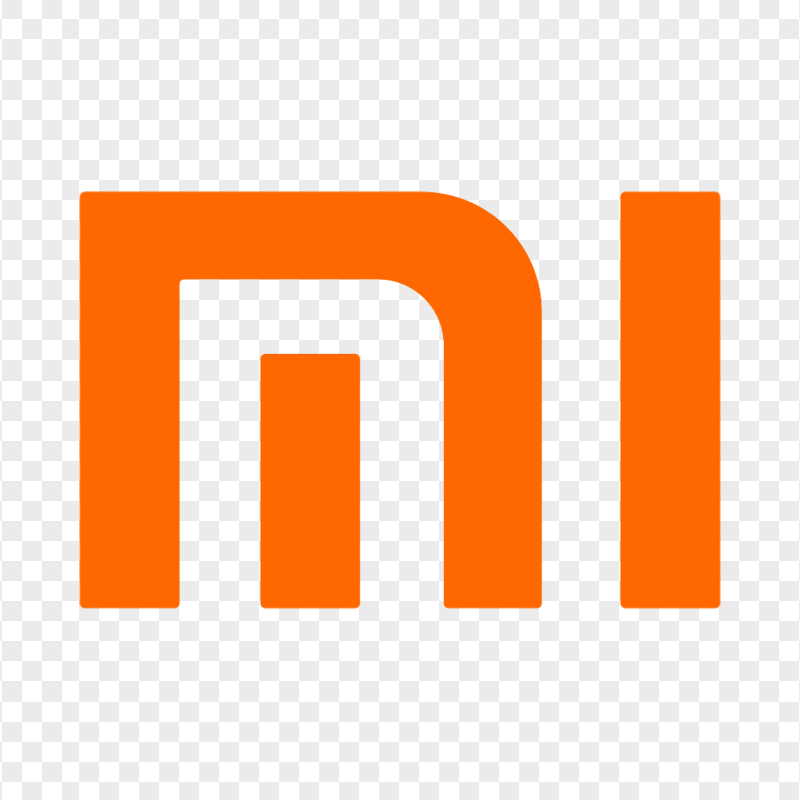}};
\node[anchor=center] at (axis cs:4, 37.7) {\includegraphics[width=0.65cm]{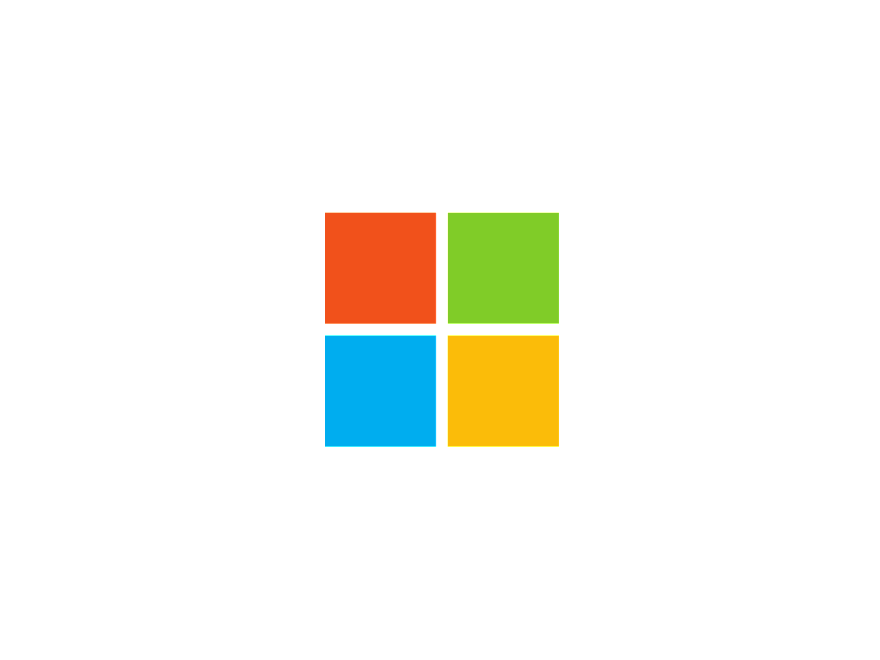}};
\node[anchor=center] at (axis cs:8, 23.6) {\includegraphics[width=0.20cm]{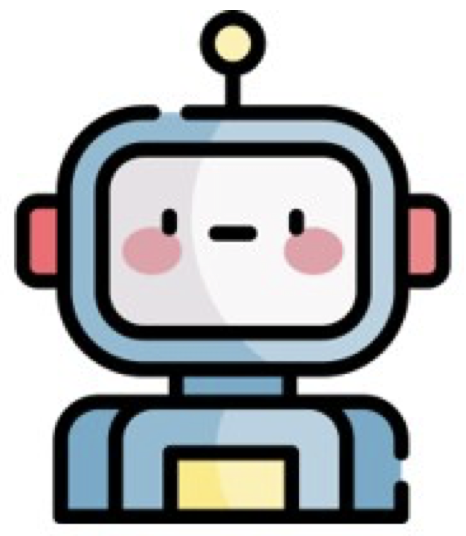}};
\node[anchor=center] at (axis cs:8, 31.4) {\includegraphics[width=\logosize]{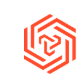}};
\node[anchor=center] at (axis cs:8, 29.2) {\includegraphics[width=0.20cm]{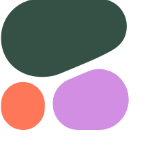}};
\node[anchor=center] at (axis cs:8, 25.9) {\includegraphics[width=\logosize]{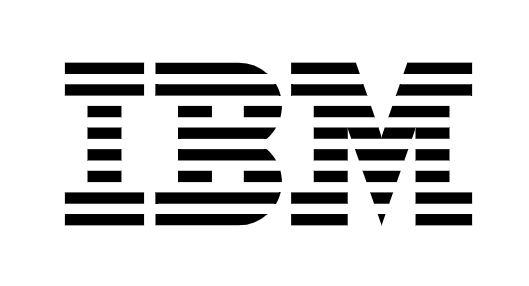}};


\node[font=\tiny\bfseries, left=0.12cm, fill=green!10, draw=green!50,rounded corners=1pt, inner sep=0.5pt] 
    at (axis cs:12,59.4) {Gemma-3 (59.4\%)};

\node[font=\tiny, right=0.12cm, inner sep=0.5pt] 
    at (axis cs:7,53.5) {Falcon-h1 (53.5\%)};
\node[font=\tiny, right=0.12cm, inner sep=0.5pt]  
    at (axis cs:4,44.9) {Gemma-3 (44.9\%)};

\node[font=\tiny, left=0.12cm, inner sep=0.5pt] 
    at (axis cs:14,44.9) {R1-Qwen};
\node[font=\tiny, right=0.12cm, inner sep=0.5pt] 
    at (axis cs:7,38.3) {R1-Qwen};
\node[font=\tiny, left=0.12cm, inner sep=0.5pt] 
    at (axis cs:4,37.7) {Phi-4};

\node[font=\tiny, right=0.12cm, inner sep=0.5pt] 
    at (axis cs:3,29.7) {Llama-3.2};
\node[font=\tiny, right=0.12cm, inner sep=0.5pt] 
    at (axis cs:1,17.0) {MobileLLM-Pro};
\node[font=\tiny, right=0.12cm, inner sep=0.5pt] 
    at (axis cs:1.5,19.6) {R1-Qwen};

\node[font=\tiny, left=0.12cm, inner sep=0.5pt] 
    at (axis cs:4,24.5) {Nemotron};

\node[font=\tiny, right=0.12cm, inner sep=0.5pt] 
    at (axis cs:7,40.2) {Qwen2.5};
\node[font=\tiny, left=0.12cm, inner sep=0.5pt] 
    at (axis cs:7,25.6) {MiMo};
\node[font=\tiny, right=0.12cm, inner sep=0.5pt] 
    at (axis cs:7,17.2) {Mistral};
\node[font=\tiny, below left=0.12cm, inner sep=0.5pt] 
    at (axis cs:7,23.7) {Olmo-3};

\node[font=\tiny, left=0.12cm, inner sep=0.5pt] 
    at (axis cs:8,42.9) {Llama-3.1};
\node[font=\tiny, right=0.12cm, inner sep=0.5pt] 
    at (axis cs:8,34.6) {Ministral};
\node[font=\tiny, right=0.12cm, inner sep=0.5pt] 
    at (axis cs:8,31.4) {Alif-1.0};
\node[font=\tiny, right=0.12cm, inner sep=0.5pt] 
    at (axis cs:8,29.2) {Aya-Expanse};
\node[font=\tiny, right=0.12cm, inner sep=0.5pt] 
    at (axis cs:8,27.6) {Rnj};
\node[font=\tiny, right=0.12cm, inner sep=0.5pt] 
    at (axis cs:8,25.9) {Granite-3.3};
\node[font=\tiny, right=0.12cm, inner sep=0.5pt] 
    at (axis cs:8,23.6) {Qalb-1.0};
\node[font=\tiny, right=0.12cm, inner sep=0.5pt] 
    at (axis cs:8,21.4) {Mistral-NeMo};

\node[font=\tiny, right=0.12cm, inner sep=0.5pt] 
    at (axis cs:9,42.2) {Gemma2};

\addplot[
    domain=1:13,
    samples=100,
    color=blue!50,
    dashed,
    line width=1.2pt,
    forget plot,
] {15 + 3.5*x};

\end{axis}
\end{tikzpicture}
\caption{Performance scaling of state-of-the-art LLMs on Urdu benchmark datasets. Average accuracy is the average of MGSM (CoT), MATH-500 (CoT), CommonSenseQA (Direct), and OpenBookQA (Direct) accuracies. Green highlights the top performing model in Urdu.}
\label{fig:model_scaling}
\end{figure}


Existing Urdu evaluation resources discussed in Section~\ref{sec:related_work} focus primarily on evaluating LLMs' capabilities in Natural Language Processing (NLP) and Natural Language Understanding (NLU) tasks that include machine translation, sentiment analysis, basic classification, or extractive question answering. While valuable, these evaluations provide only a partial view of model capability and do not assess complex reasoning such as mathematical problem solving or common sense reasoning. In addition, regional initiatives, including TituLLM \cite{nahin2025titullmsfamilybanglallms} for Bangla, C²RBench \cite{cbench} for Chinese, \cite{arabicsense} , \cite{arareasoner} for Arabic, and MasakhaNER for African languages \cite{adelani2021masakhaner} further highlight local efforts to benchmark complex model behavior in various linguistic settings. 

To address this gap, we introduce a reasoning-focused benchmark for evaluating LLMs in Urdu by translating four widely adopted datasets including Multilingual Grade School Math (MGSM) \cite{mgsm}, OpenBookQA \cite{obqa}, CommonSenseQA \cite{csqa} and MATH-500 \cite{math500} into high-quality Urdu. These datasets span diverse and complementary reasoning abilities that have not been systematically evaluated in Urdu. MGSM measures mathematical and multi-step reasoning, OpenBookQA assesses factual and retrieval-based reasoning, and CommonSenseQA evaluates commonsense knowledge and inference.  

To ensure translation fidelity, we employ a contextual ensemble translation framework with human-in-the-loop validation as examples reported in Table~\ref{tab:one_example_context} and Appendix~\ref{appendix:context-transaltion}. Using the resulting datasets, we benchmark multiple state-of-the-art reasoning and non-reasoning LLMs under a unified LM Evaluation Harness protocol, enabling large-scale and standardized comparison of reasoning performance in Urdu as shown in Figure~\ref{fig:model_scaling}.

\paragraph{Key Findings}
The key findings of this study are summarized below:

\begin{enumerate}
    \item \textbf{Scalable Urdu Translation Framework (\S\ref{sec:contextual_translations})} 
    A contextual ensemble translation pipeline with human-in-the-loop validation produces high-fidelity Urdu translations of MGSM, MATH-500, CommonSenseQA, and OpenBookQA, enabling reproducible evaluation.

    \item \textbf{Comprehensive Model Evaluations (\S\ref{sec:results})} 
    We evaluate diverse reasoning and non-reasoning LLMs under multiple prompting strategies in a unified, reproducible setup.

    \item \textbf{Dataset Sensitivity (\S\ref{sec:dataset_impact})} 
    Reasoning benchmarks (MGSM, MATH-500) show higher variance than commonsense tasks (CommonSenseQA, OpenBookQA), reflecting increased symbolic and multi-step reasoning difficulty.

    \item \textbf{Impact of Task Difficulty (\S\ref{sec:task_impact})} 
    MATH-500 accuracy declines monotonically with difficulty; non-instruction-tuned models degrade sharply, while DeepSeek-R1-Distill-Qwen-14B and Phi-4-Mini-Reasoning better sustain performance.

    \item \textbf{Scaling Effects (\S\ref{sec:model_scaling_impact})} 
    Larger models (Gemma-3-12B-it, Falcon-h1-7B-Instruct) generally perform better, but scale alone is insufficient, as Gemma-3-4B-it outperforms several 7B models.

    \item \textbf{Reasoning vs. Instruction Tuning (\S\ref{sec:reasoning_impact})} 
    Reasoning-oriented models improve arithmetic reasoning on MGSM but do not consistently outperform instruction-tuned models on commonsense benchmarks.

    \item \textbf{Language Consistency Test (\S\ref{sec:language_impact})} 
    Higher Urdu language consistency correlates with improved reasoning accuracy, particularly on MGSM under Chain-of-Thought prompting, where Gemma, Llama, and Qwen show near-complete Urdu adherence.
\end{enumerate}

\begin{figure*}[t]
    \centering
    \includegraphics[width=0.8\linewidth]{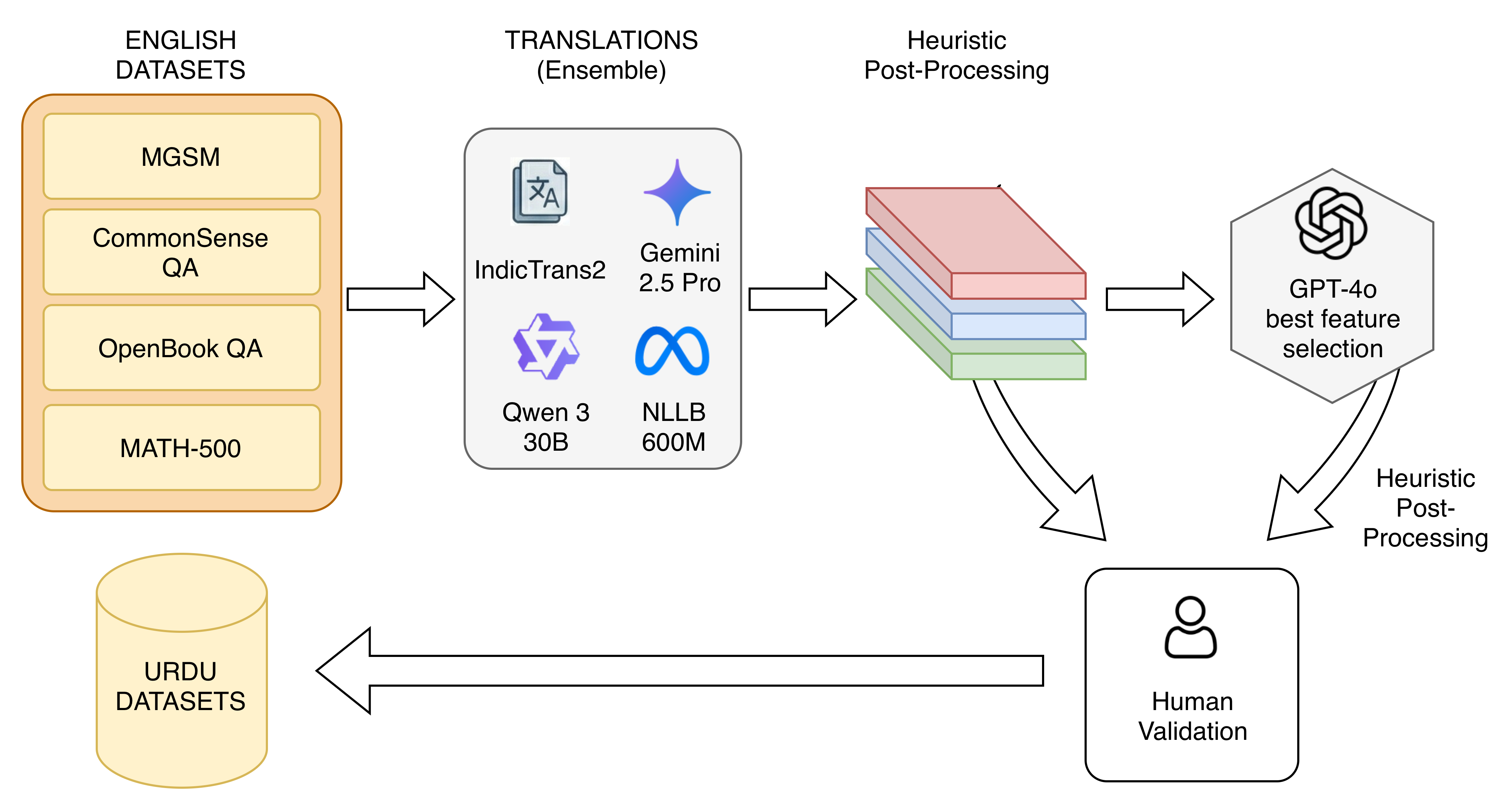}
    \caption{Contextual Ensembled Translations Framework with human-in-the-loop}
    \label{fig:flowchart}
\end{figure*}

\section{Related Work}
\label{sec:related_work}
Prior work shows that automatically translated benchmarks, including Spanish MMLU, introduce substantial errors that distort evaluation outcomes \cite{plaza2024spanishllmbenchmarksmmlu, 11232191}. Large-scale multilingual studies across European languages further confirm strong sensitivity to translation quality and services \cite{thellmann2024multilingualllmevaluationeuropean}. Machine translation research highlights the limitations of existing evaluation metrics and the instability of LLM-based translation, particularly for low-resource languages \cite{lópezcaro2023, zhu-etal-2024-multilingual}. Recent benchmarks for Bengali and other underrepresented languages similarly reveal persistent performance gaps and the necessity of human validation for reliable evaluation \cite{bhowmik2025evaluatingllmsmultilingualcapabilities, song2025smalllanguagemodelsilver}. In contrast, we introduce a scalable, context-aware ensemble translation framework with human-in-the-loop validation for reliable Urdu reasoning evaluation.

\paragraph{Urdu Benchmarking Datasets} Prior Urdu evaluations primarily target general NLP capabilities rather than structured reasoning. Existing work compares generalist and specialist models across Urdu tasks \cite{arif2024generalists}, studies cross-lingual QA and translation-induced errors \cite{kazi-etal-2025-crossing}, and probes linguistic competence using minimal-pair benchmarks such as UrBLiMP \cite{arshad2024urb}. Other efforts address factual verification \cite{ahmad-etal-2025-urdufactcheck} or classical NLP tasks. However, these benchmarks do not systematically evaluate multi-step mathematical or commonsense reasoning. UrduBench addresses this gap by providing standardized, context-preserving, and human-validated Urdu translations of established reasoning benchmarks.

\section{Contextually Ensembled Translation Method}
\label{sec:contextual_translations}



Constructing reliable reasoning benchmarks in Urdu requires more than direct machine-translation, particularly for structured and multi-step reasoning tasks. We observe that multiple-choice datasets such as OpenBookQA and CommonSenseQA are especially prone to contextual fragmentation when questions and answer options are translated independently, leading to semantically misaligned Urdu outputs. To mitigate this, we employ a contextually ensembled, human-in-the-loop translation pipeline that preserves contextual coherence while integrating multiple translation models and quality controls as shown in Figure~\ref{fig:flowchart}. The examples in Appendix \ref{appendix:context-transaltion} also illustrate how the context-aware translation strategy improves translation quality.

\subsection{Contextual Translations}

We translate MGSM, CommonSenseQA, OpenBookQA, and MATH-500 into Urdu using a unified pipeline designed to preserve semantic fidelity, contextual coherence, and structural integrity. Each English instance is first translated independently using four complementary translation candidates: IndicTrans2~\cite{gala2023indictrans2highqualityaccessiblemachine}, NLLB~\cite{nllb} , Qwen-3-30B~\cite{yang2025qwen3technicalreport}, and Gemini-2.5-Pro~\cite{comanici2025gemini25pushingfrontier}. These candidates are selected based on their demonstrated translation quality scores in prior multilingual evaluation studies~\cite{basit-etal-2024-challenges, nahin2025titullmsfamilybanglallms}. Translation prompts used for these candidates are provided in Appendix~\ref{a1}.

For structured multiple-choice datasets such as CommonSenseQA and OpenBookQA, we adopt a context-preserving translation strategy by concatenating the question stem with all associated answer choices prior to translation. This prevents semantic fragmentation and ensures grammatical and logical alignment between the question and its answer options in Urdu. After translation, questions and answer choices are programmatically separated into their original components.

\subsection{Candidate Translation Fusion}

GPT-5.1~\cite{openai} is used to refine candidate translations by leveraging complementary strengths across models in lexical choice, grammar, and handling of technical content. It compares candidates proposed translations against the original English source to either synthesize an improved Urdu translation or select the best-performing output. Beyond quality improvement, this stage also serves as an automated judging step and limiting the need for manual human validation. The prompt for best feature selection and representative examples are provided in Appendix~\ref{a2}.

\subsection{Heuristic Post-Processing}

All translations undergo automatic heuristic validation tailored to Urdu, as described in Algorithm~\ref{alg:urdu-quality}. The validation detects the following issues in translated outputs:
\begin{itemize}
  \item Empty or incomplete translations
  \item Missing Urdu questions or answer options
  \item Abnormally short translated questions
  \item Invalid or missing answer keys
  \item Excessive residual English content
\end{itemize}

Most translations pass these checks automatically; flagged cases are reviewed and reprocessed to ensure accurate and complete translations.

\subsection{Human Validation}

All candidate translations, including those produced by GPT-5.1, undergo final review by professional native Urdu-speaking annotators fluent in English. Annotators\footnote{Paid hourly; above min. wage of country of employment.} follow detailed guidelines provided in Appendix~\ref{appendix:human-guidlines} and select the most accurate and natural translation for each instance. This step is essential for resolving residual ambiguities, ensuring semantic fidelity, and accounting for cultural and linguistic nuances that automated systems may overlook. Using this framework, the best quality translation is selected for inclusion in the final benchmark, ensuring linguistic coherence and contextual fidelity suitable for evaluating LLM reasoning in Urdu. Dataset statistics for the resulting benchmarks are reported in Table~\ref{tab:dataset-stats}. Human Validation Win--Loss Statistics Across Translation Candidates in Table~\ref{tab:human_win_loss} while human preference win--loss comparison between GPT-5.1 refined and human-edited Urdu translations across four datasets in Figure~\ref{fig:human_win_loss} in Appendix~\ref{a4}.

\begin{table}[t]
\centering
\small
\renewcommand{\arraystretch}{1.4}
\begin{tabular}{l|c|c|c}
\hline
\textbf{Dataset} & \textbf{Train} & \textbf{Val} & \textbf{Test} \\
\hline
MGSM            & 8    & --   & 250  \\
CommonSenseQA   & 9{,}741 & 1{,}221 & 1{,}140 \\
OpenBookQA      & 4{,}957 & 500 & 500 \\
MATH-500        & --   & --   & 500 \\
\hline
\end{tabular}
\caption{Statistics of the translated Urdu datasets.}
\label{tab:dataset-stats}
\end{table}

\begin{table*}[h!]
\centering
\hspace*{-2cm}
\small
\renewcommand{\arraystretch}{1.4}
\caption{Evaluation Across All Urdu Benchmark Datasets. Metric: Normalized Accuracy }
\renewcommand{\arraystretch}{2.0}
\setlength{\tabcolsep}{4pt}
\begin{tabular}{l|ccc|c|c|c|c}
\hline
\textbf{Model} & \multicolumn{3}{c|}{\textbf{MGSM}} & \textbf{CSQA} & \textbf{OBQA} & \textbf{M500} & \textbf{Avg} \\
 & \textbf{Direct} & \textbf{CoT} & \textbf{FS+CoT} & \textbf{Direct} & \textbf{Direct} & \textbf{CoT} & \textbf{\%} \\ 
\hline
\hline \multicolumn{8}{c}{\textit{Non-Reasoning-oriented Open-source LLM}} \\  \hline 
MobileLLM-Pro (1B) & 0.40\scriptsize{$\pm$0.0} & 0.00\scriptsize{$\pm$0.0} & 0.40\scriptsize{$\pm$0.4} & 20.8\scriptsize{$\pm$1.2} & 27.4\scriptsize{$\pm$2.0} & 2.80\scriptsize{$\pm$0.8} & 17.0\scriptsize{$\pm$1.0} \\ 

Llama-3.2-3B-Instruct & 28.4\scriptsize{$\pm$2.9} & 40.0\scriptsize{$\pm$3.0} & 35.6\scriptsize{$\pm$3.0} & 37.1\scriptsize{$\pm$1.4} & 29.4\scriptsize{$\pm$2.0} & 12.6\scriptsize{$\pm$1.5} & 29.7\scriptsize{$\pm$1.9} \\

Gemma-3-4B-it & 54.8\scriptsize{$\pm$3.2} & 64.0\scriptsize{$\pm$3.0} & 44.0\scriptsize{$\pm$3.2} & 40.5\scriptsize{$\pm$1.4} & 31.2\scriptsize{$\pm$2.1} & 44.2\scriptsize{$\pm$2.2} & 44.9\scriptsize{$\pm$2.2} \\ 

Nemotron-Mini-4B & 18.8\scriptsize{$\pm$2.5} & 22.4\scriptsize{$\pm$2.6} & 25.6\scriptsize{$\pm$2.8} & 38.4\scriptsize{$\pm$1.4} & 32.0\scriptsize{$\pm$2.1} & 5.00\scriptsize{$\pm$1.0} & 24.5\scriptsize{$\pm$1.8} \\

Falcon-h1-7b-instruct & 71.2\scriptsize{$\pm$2.9} & 72.4\scriptsize{$\pm$2.8} & 70.4\scriptsize{$\pm$2.9} & \cellcolor{green!20} 59.3\scriptsize{$\pm$1.4} & 32.8\scriptsize{$\pm$2.1} &   49.6\scriptsize{$\pm$2.2} & 53.5\scriptsize{$\pm$2.1} \\ 

MiMo-7B-Base & 29.6\scriptsize{$\pm$2.9} & 20.8\scriptsize{$\pm$2.6} & 56.4\scriptsize{$\pm$3.1} & 37.4\scriptsize{$\pm$1.4} & 28.8\scriptsize{$\pm$2.0} & 15.4\scriptsize{$\pm$1.6} & 25.6\scriptsize{$\pm$2.2} \\

Mistral-7B-Instruct & 0.40\scriptsize{$\pm$1.2} & 4.40\scriptsize{$\pm$1.3} & 11.6\scriptsize{$\pm$2.0} & 27.5\scriptsize{$\pm$1.3} & 30.2\scriptsize{$\pm$2.1} & 6.60\scriptsize{$\pm$1.1} & 17.2\scriptsize{$\pm$1.5} \\

Olmo-3-7B-Instruct & 21.6\scriptsize{$\pm$2.6} & 23.6\scriptsize{$\pm$2.7} & 24.0\scriptsize{$\pm$2.7} & 27.9\scriptsize{$\pm$1.3} & 26.6\scriptsize{$\pm$2.0} & 20.6\scriptsize{$\pm$1.8} & 24.7\scriptsize{$\pm$1.9} \\ 

Qwen2.5-7B-Instruct & 53.2\scriptsize{$\pm$3.2} & 57.6\scriptsize{$\pm$3.1} & 60.4\scriptsize{$\pm$3.1} & 30.4\scriptsize{$\pm$1.3} & 31.6\scriptsize{$\pm$2.1} & 41.2\scriptsize{$\pm$2.2} & 40.2\scriptsize{$\pm$2.2} \\ 

Alif-1.0-8B-Instruct & 21.6\scriptsize{$\pm$2.6} & 30.4\scriptsize{$\pm$2.9} & 43.6\scriptsize{$\pm$3.1} & 51.3\scriptsize{$\pm$1.4} & 33.4\scriptsize{$\pm$2.1} & 10.6\scriptsize{$\pm$1.4} & 31.4\scriptsize{$\pm$1.9} \\ 

Aya-Expanse-8B & 22.0\scriptsize{$\pm$2.6} & 30.8\scriptsize{$\pm$2.9} & 28.8\scriptsize{$\pm$2.8} & 37.5\scriptsize{$\pm$1.4} & 31.6\scriptsize{$\pm$2.1} & 17.2\scriptsize{$\pm$1.7} & 29.2\scriptsize{$\pm$2.0} \\ 

Granite-3.3-8b-instruct & 8.00\scriptsize{$\pm$1.7} & 33.6\scriptsize{$\pm$3.0} & 30.4\scriptsize{$\pm$2.9} & 28.2\scriptsize{$\pm$1.3} & 29.6\scriptsize{$\pm$0.0} & 12.4\scriptsize{$\pm$1.5} & 25.9\scriptsize{$\pm$1.4} \\

Qalb-1.0-8B-Instruct & 19.6\scriptsize{$\pm$2.5} & 17.6\scriptsize{$\pm$2.4} & 7.60\scriptsize{$\pm$1.7} & 38.4\scriptsize{$\pm$1.4} & 32.2\scriptsize{$\pm$2.1} & 6.00\scriptsize{$\pm$1.1} & 23.6\scriptsize{$\pm$1.6} \\

Llama-3.1-8B-Instruct & 62.0\scriptsize{$\pm$3.1} & 62.8\scriptsize{$\pm$3.1} & 54.4\scriptsize{$\pm$3.2} & 52.8\scriptsize{$\pm$1.4} & 33.8\scriptsize{$\pm$2.1} & 22.4\scriptsize{$\pm$1.9} & 42.9\scriptsize{$\pm$1.7} \\ 

Ministral-8B-Instruct-2410 & 55.2\scriptsize{$\pm$3.2} & 58.4\scriptsize{$\pm$3.1} & 50.8\scriptsize{$\pm$3.2} & 29.6\scriptsize{$\pm$1.3} & 30.8\scriptsize{$\pm$2.1} & 19.4\scriptsize{$\pm$1.8} & 34.6\scriptsize{$\pm$2.1} \\ 

Mistral-NeMo-Minitron-8B-Instruct & 4.00\scriptsize{$\pm$1.2} & 0.40\scriptsize{$\pm$0.4} & 0.40\scriptsize{$\pm$0.4} & 39.2\scriptsize{$\pm$1.4} & 29.0\scriptsize{$\pm$2.0} & 17.0\scriptsize{$\pm$1.7} & 21.4\scriptsize{$\pm$1.4} \\ 
 
Rnj-1-instruct (8B) & 22.4\scriptsize{$\pm$2.6} & 13.2\scriptsize{$\pm$2.2} & 9.20\scriptsize{$\pm$1.8} & 27.9\scriptsize{$\pm$1.3} & 26.6\scriptsize{$\pm$2.0} & 42.6\scriptsize{$\pm$2.2} & 27.6\scriptsize{$\pm$1.9} \\

Gemma2-9B-it & 74.0\scriptsize{$\pm$2.8} & 73.6\scriptsize{$\pm$2.8} & 75.2\scriptsize{$\pm$2.7} & 25.3\scriptsize{$\pm$1.2} & 32.8\scriptsize{$\pm$2.1} & 37.0\scriptsize{$\pm$2.2} & 42.2\scriptsize{$\pm$2.1} \\ 

Gemma-3-12B-it & \cellcolor{green!20} 80.4\scriptsize{$\pm$2.5} & \cellcolor{green!20} 80.8\scriptsize{$\pm$2.5} & \cellcolor{green!20} 82.8\scriptsize{$\pm$2.4} & 49.4\scriptsize{$\pm$1.4} & \cellcolor{green!20} 37.4\scriptsize{$\pm$2.2} & \cellcolor{green!20} 70.0\scriptsize{$\pm$2.1} & \cellcolor{green!20} 59.4\scriptsize{$\pm$2.1} \\ 

\hline

\multicolumn{8}{c}{\textit{Reasoning-oriented Open-source LLM}} \\ \hline

DeepSeek-R1-Distill-Qwen-1.5B & 8.40\scriptsize{$\pm$1.8} & 6.40\scriptsize{$\pm$1.6} & 9.60\scriptsize{$\pm$1.9} & 20.6\scriptsize{$\pm$1.2} & 30.0\scriptsize{$\pm$2.1} & 21.4\scriptsize{$\pm$1.8} & 19.6\scriptsize{$\pm$1.7} \\

Phi-4-mini-reasoning (4B) & 38.4\scriptsize{$\pm$3.1} & 32.4\scriptsize{$\pm$2.9} & 29.2\scriptsize{$\pm$2.9} & 20.4\scriptsize{$\pm$1.2} & 32.4\scriptsize{$\pm$2.1} & 65.8\scriptsize{$\pm$2.1} & 37.7\scriptsize{$\pm$2.1} \\ 

DeepSeek-R1-Distill-Qwen-7B & 39.6\scriptsize{$\pm$3.1} & 42.0\scriptsize{$\pm$3.1} & 36.0\scriptsize{$\pm$3.0} & 22.9\scriptsize{$\pm$1.2} & 31.2\scriptsize{$\pm$2.1} & 57.0\scriptsize{$\pm$2.2} & 38.3\scriptsize{$\pm$2.2} \\

DeepSeek-R1-Distill-Qwen-14B & 54.4\scriptsize{$\pm$3.2} & 59.2\scriptsize{$\pm$3.1} & 68.8\scriptsize{$\pm$2.9} & 19.4\scriptsize{$\pm$1.1} & 31.8\scriptsize{$\pm$2.1} & 69.2\scriptsize{$\pm$2.1} & 44.9\scriptsize{$\pm$1.2} \\ 

\hline
\end{tabular}

\label{tab:eval-formats}
\footnotesize\begin{flushleft}
~~~~~~~~CSQA:  CommonSenseQA.\\
~~~~~~~~OBQA: OpenBookQA.\\
~~~~~~~~Average considers MGSM (cot), MATH-500 (cot) CSQA (direct), and OBQA (direct).
\end{flushleft}

\end{table*}

\section{Evaluation of Models}

In this section, we present a systematic evaluation of diverse language models on the proposed Urdu benchmarks, analyzing the effects of model design, prompting strategy, and task difficulty on reasoning performance.

\subsection{Experimental Setup and Datasets}

We evaluate models on an Urdu-translated reasoning and question-answering benchmarks introduced in this work: 
\textbf{MGSM} (arithmetic reasoning, evaluated with Direct, CoT, and three Few-Shot + CoT prompting), 
\textbf{MATH-500} (symbolic mathematics across five difficulty levels, L1--L5), 
\textbf{CommonSenseQA} (commonsense reasoning), and 
\textbf{OpenBookQA} (knowledge-based scientific reasoning). 
Together, these benchmarks span arithmetic, formal mathematics, commonsense understanding, and factual knowledge translated in Urdu.



\subsection{Evaluated Models}

We evaluate a broad set of open-source and open-weight LLMs, including instruction-tuned and explicitly reasoning-oriented models, under a fixed evaluation protocol as shown in Table~\ref{tab:eval-formats}. The models span parameter scales from 1B to 14B+, diverse multilingual pretraining regimes, and different reasoning recipes, including distillation- and RL-based approaches (e.g., DeepSeek-R1, Phi-4-Mini-Reasoning) as shown in Table~\ref{tab:model-size-categories}. This diversity enables controlled analysis of how various model architectures, model scaling, training strategy, prompting methods, and language consistency, influence Urdu reasoning performance, particularly under increasing task difficulty as demonstrated in Figure~\ref{fig:model_performance_levels}. All evaluation prompts, configurations and model details are provided in Appendix~\ref{app:prompts}.

\subsection{Results}
\label{sec:results}


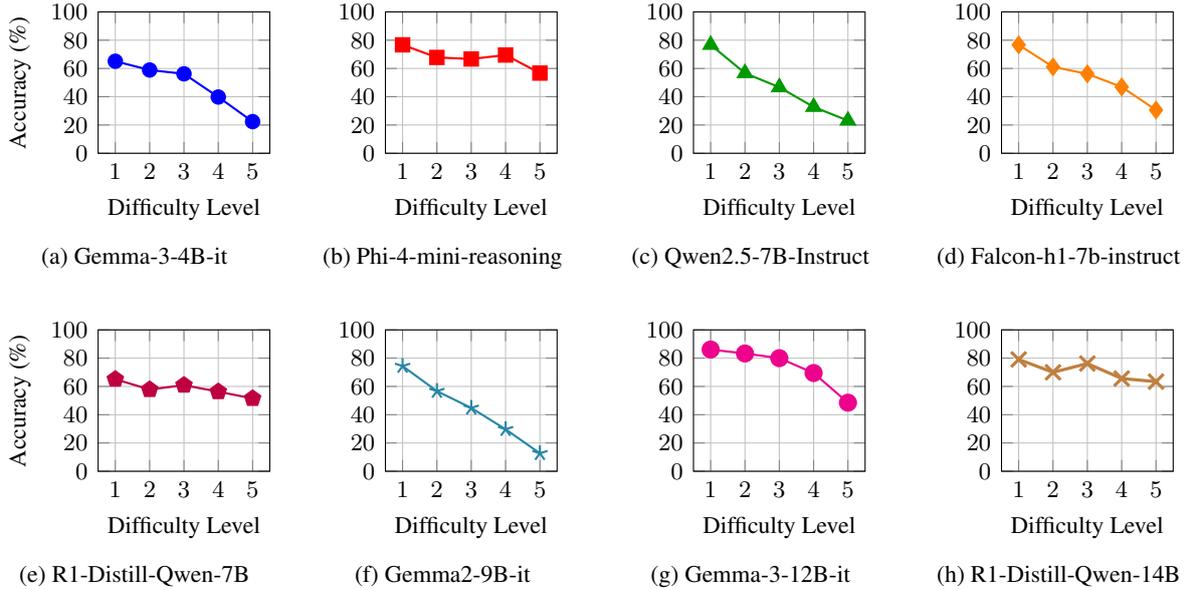
\begin{figure*}[t]
\centering

\begin{subfigure}[b]{0.24\textwidth}
\centering
\begin{tikzpicture}
\begin{axis}[
    width=\textwidth,
    height=0.9\textwidth,
    xlabel={Difficulty Level},
    ylabel={Accuracy (\%)},
    xmin=0.5, xmax=5.5,
    ymin=0, ymax=100,
    xtick={1,2,3,4,5},
    ytick={0,20,40,60,80,100},
    tick label style={font=\small},
    label style={font=\small},
    grid=major,
    grid style={line width=.1pt, draw=gray!30},
    major grid style={line width=.2pt, draw=gray!50},
]
\addplot[color=blue, mark=*, thick, mark size=2.5pt] coordinates {
    (1,65.1) (2,58.9) (3,56.2) (4,39.8) (5,22.4)
};
\end{axis}
\end{tikzpicture}
\caption{Gemma-3-4B-it}
\label{fig:gemma3-4b}
\end{subfigure}
\hfill
\begin{subfigure}[b]{0.24\textwidth}
\centering
\begin{tikzpicture}
\begin{axis}[
    width=\textwidth,
    height=0.9\textwidth,
    xlabel={Difficulty Level},
    xmin=0.5, xmax=5.5,
    ymin=0, ymax=100,
    xtick={1,2,3,4,5},
    ytick={0,20,40,60,80,100},
    tick label style={font=\small},
    label style={font=\small},
    grid=major,
    grid style={line width=.1pt, draw=gray!30},
    major grid style={line width=.2pt, draw=gray!50},
]
\addplot[color=red, mark=square*, thick, mark size=2.5pt] coordinates {
    (1,76.7) (2,67.8) (3,66.7) (4,69.5) (5,56.72)
};
\end{axis}
\end{tikzpicture}
\caption{Phi-4-mini-reasoning}
\label{fig:phi4}
\end{subfigure}
\hfill
\begin{subfigure}[b]{0.24\textwidth}
\centering
\begin{tikzpicture}
\begin{axis}[
    width=\textwidth,
    height=0.9\textwidth,
    xlabel={Difficulty Level},
    xmin=0.5, xmax=5.5,
    ymin=0, ymax=100,
    xtick={1,2,3,4,5},
    ytick={0,20,40,60,80,100},
    tick label style={font=\small},
    label style={font=\small},
    grid=major,
    grid style={line width=.1pt, draw=gray!30},
    major grid style={line width=.2pt, draw=gray!50},
]
\addplot[color=green!60!black, mark=triangle*, thick, mark size=3pt] coordinates {
    (1,76.7) (2,56.7) (3,46.7) (4,32.8) (5,23.1)
};
\end{axis}
\end{tikzpicture}
\caption{Qwen2.5-7B-Instruct}
\label{fig:qwen}
\end{subfigure}
\hfill
\begin{subfigure}[b]{0.24\textwidth}
\centering
\begin{tikzpicture}
\begin{axis}[
    width=\textwidth,
    height=0.9\textwidth,
    xlabel={Difficulty Level},
    xmin=0.5, xmax=5.5,
    ymin=0, ymax=100,
    xtick={1,2,3,4,5},
    ytick={0,20,40,60,80,100},
    tick label style={font=\small},
    label style={font=\small},
    grid=major,
    grid style={line width=.1pt, draw=gray!30},
    major grid style={line width=.2pt, draw=gray!50},
]
\addplot[color=orange, mark=diamond*, thick, mark size=3pt] coordinates {
    (1,76.7) (2,61.1) (3,56.2) (4,46.9) (5,30.6)
};
\end{axis}
\end{tikzpicture}
\caption{Falcon-h1-7b-instruct}
\label{fig:falcon}
\end{subfigure}

\vspace{0.5cm}

\begin{subfigure}[b]{0.24\textwidth}
\centering
\begin{tikzpicture}
\begin{axis}[
    width=\textwidth,
    height=0.9\textwidth,
    xlabel={Difficulty Level},
    ylabel={Accuracy (\%)},
    xmin=0.5, xmax=5.5,
    ymin=0, ymax=100,
    xtick={1,2,3,4,5},
    ytick={0,20,40,60,80,100},
    tick label style={font=\small},
    label style={font=\small},
    grid=major,
    grid style={line width=.1pt, draw=gray!30},
    major grid style={line width=.2pt, draw=gray!50},
]
\addplot[color=purple, mark=pentagon*, thick, mark size=3pt] coordinates {
    (1,65.1) (2,57.8) (3,60.9) (4,56.3) (5,51.5)
};
\end{axis}
\end{tikzpicture}
\caption{R1-Distill-Qwen-7B}
\label{fig:r1-7b}
\end{subfigure}
\hfill
\begin{subfigure}[b]{0.24\textwidth}
\centering
\begin{tikzpicture}
\begin{axis}[
    width=\textwidth,
    height=0.9\textwidth,
    xlabel={Difficulty Level},
    xmin=0.5, xmax=5.5,
    ymin=0, ymax=100,
    xtick={1,2,3,4,5},
    ytick={0,20,40,60,80,100},
    tick label style={font=\small},
    label style={font=\small},
    grid=major,
    grid style={line width=.1pt, draw=gray!30},
    major grid style={line width=.2pt, draw=gray!50},
]
\addplot[color=cyan!60!black, mark=star, thick, mark size=3pt] coordinates {
    (1,74.4) (2,56.7) (3,44.8) (4,29.7) (5,12.7)
};
\end{axis}
\end{tikzpicture}
\caption{Gemma2-9B-it}
\label{fig:gemma2-9b}
\end{subfigure}
\hfill
\begin{subfigure}[b]{0.24\textwidth}
\centering
\begin{tikzpicture}
\begin{axis}[
    width=\textwidth,
    height=0.9\textwidth,
    xlabel={Difficulty Level},
    xmin=0.5, xmax=5.5,
    ymin=0, ymax=100,
    xtick={1,2,3,4,5},
    ytick={0,20,40,60,80,100},
    tick label style={font=\small},
    label style={font=\small},
    grid=major,
    grid style={line width=.1pt, draw=gray!30},
    major grid style={line width=.2pt, draw=gray!50},
]
\addplot[color=magenta, mark=otimes*, thick, mark size=3pt] coordinates {
    (1,86.1) (2,83.3) (3,80.0) (4,69.5) (5,48.5)
};
\end{axis}
\end{tikzpicture}
\caption{Gemma-3-12B-it}
\label{fig:gemma3-12b}
\end{subfigure}
\hfill
\begin{subfigure}[b]{0.24\textwidth}
\centering
\begin{tikzpicture}
\begin{axis}[
    width=\textwidth,
    height=0.9\textwidth,
    xlabel={Difficulty Level},
    xmin=0.5, xmax=5.5,
    ymin=0, ymax=100,
    xtick={1,2,3,4,5},
    ytick={0,20,40,60,80,100},
    tick label style={font=\small},
    label style={font=\small},
    grid=major,
    grid style={line width=.1pt, draw=gray!30},
    major grid style={line width=.2pt, draw=gray!50},
]
\addplot[color=brown, mark=x, thick, mark size=4pt, line width=1.2pt] coordinates {
    (1,79.1) (2,70.0) (3,76.2) (4,65.6) (5,63.4)
};
\end{axis}
\end{tikzpicture}
\caption{R1-Distill-Qwen-14B}
\label{fig:r1-14b}
\end{subfigure}

\caption{Performance of different language models across difficulty levels (L1--L5) on the MATH-500-Urdu benchmark. Each subplot shows the accuracy degradation pattern as problem difficulty increases.}
\label{fig:model_performance_levels}
\end{figure*}

We report normalized accuracy across all Urdu benchmarks in Table~\ref{tab:eval-formats}, covering arithmetic reasoning (MGSM, MATH-500) and commonsense question answering (CommonSenseQA, OpenBookQA). Results are aggregated across multiple prompting strategies to analyze prompt sensitivity, architectural effects, and overall model robustness in a low-resource language setting.
\paragraph{Prompt Sensitivity}
Prompting strategy is a major determinant of performance on MGSM. Direct prompting consistently underperforms reasoning-aware prompts, indicating that most models struggle to implicitly perform multi-step reasoning in Urdu without explicit guidance. Introducing Chain-of-Thought (CoT) leads to substantial gains across nearly all architectures, often yielding large relative improvements over Direct prompting. However, the benefits of Few-Shot + CoT are uneven: while it enables peak performance for some models (e.g., Falcon-h1-7B-Instruct), it introduces instability or regressions for others, particularly smaller or less multilingual models. This suggests that Urdu exemplars can impose additional distributional shift rather than uniformly improving generalization. Overall, CoT emerges as the most reliable prompting strategy, while few-shot gains remain highly model-dependent.
\paragraph{Model Architecture}
Model architecture and training objectives strongly influence Urdu reasoning performance. Models with stronger multilingual pretraining and instruction tuning exhibit more reliable transfer across tasks. Gemma-family models, particularly Gemma-3-12B-it and Gemma-3-4B-it, achieve strong performance across both reasoning-intensive and commonsense benchmarks, indicating effective cross-lingual abstraction. In contrast, compact or efficiency-oriented architectures (e.g., MobileLLM-Pro, Mistral-NeMo-Minitron) perform near chance on MGSM and degrade sharply on MATH-500, reflecting limited capacity for sustained symbolic reasoning in Urdu. Reasoning-oriented distilled models (e.g., DeepSeek-R1 variants) show moderate but inconsistent improvements, suggesting that reasoning distillation alone does not fully compensate for limited multilingual grounding.
\paragraph{Overall Performance Across Benchmarks.}
Despite improvements from prompting and architecture, performance in Urdu remains uneven across tasks, underscoring the challenges of low-resource evaluation. Nevertheless, several models achieve non-trivial accuracy across benchmarks, validating both the semantic fidelity of the translated datasets and the feasibility of standardized Urdu evaluation. Gemma-3-12B-it and Falcon-h1-7B-Instruct achieve the highest overall averages, demonstrating that strong instruction tuning combined with multilingual exposure transfers most effectively. Taken together, the results indicate that reliable Urdu reasoning requires a combination of explicit reasoning prompts, sufficient model capacity, and multilingual-aware training, with no single factor being sufficient on its own.

\section{Analyses and Discussion}
In this section, we analyze the results to isolate how dataset properties, task difficulty, model scale, reasoning specialization, and language consistency influence Urdu reasoning performance.


\subsection{Impact of Dataset}
\label{sec:dataset_impact}
\textit{How does dataset choice affect model performance in Urdu?}

Dataset choice has a strong, task-dependent effect on model performance in Urdu. Reasoning-intensive benchmarks such as MGSM and MATH-500 exhibit the highest variance across models, indicating that multi-step and symbolic reasoning is particularly sensitive to linguistic transfer and translation noise in a low-resource setting as shown in Figure~\ref{fig:datasets_box_plot} in Appendix~\ref{app:extended_results}. In contrast, commonsense and knowledge-based tasks such as CommonSenseQA and OpenBookQA remain comparatively stable, with tighter performance distributions. Overall, these results indicate that reasoning depth and symbolic complexity are primary drivers of evaluation difficulty in Urdu, while semantic and knowledge-oriented tasks are more robust to cross-lingual transfer.

\subsection{Impact of Task Difficulty}
\label{sec:task_impact}
\textit{How does task difficulty affect mathematical reasoning in Urdu?}

Task difficulty has a systematic impact on performance in MATH-500-Urdu. As shown in Figure~\ref{fig:model_performance_levels} and Table~\ref{tab:math500-levels}, accuracy declines steadily with increasing problem difficulty, with the largest drops at higher levels (L4--L5). While most models perform well on low-difficulty problems, they degrade sharply on more complex subsets, indicating limited robustness to deep symbolic and multi-step reasoning in Urdu. Larger and reasoning-oriented models such as Gemma-3-12B-it and DeepSeek-R1-Distill-Qwen-14B retain comparatively stronger performance at higher difficulty levels, but the overall trend shows that linguistic transfer amplifies the inherent challenges of formal mathematics.

\subsection{Impact of Model Size}
\label{sec:model_scaling_impact}
\textit{How does model scale influence performance in Urdu?}

Model scale contributes to performance on Urdu benchmarks, but it is not sufficient on its own. While larger models such as Gemma-3-12B-it achieve strong overall results, scale does not guarantee superiority. Notably, Gemma-3-4B-it outperforms several larger models, including Llama-3.1-8B-Instruct and DeepSeek-R1-Distill-Qwen-7B as shown in Table~\ref{tab:eval-formats} and in Figure~\ref{fig:model_scaling_2}, highlighting the importance of multilingual training and Urdu exposure. Similarly, Gemma-3-12B-it surpasses the larger DeepSeek-R1-Distill-Qwen-14B on reasoning-intensive benchmarks, indicating that reasoning-oriented scaling alone does not ensure better low-resource language performance. Overall, these results show that model size is only one factor, with targeted multilingual coverage playing a critical role in Urdu reasoning quality.

\subsection{Impact of Reasoning and Non-Reasoning Models}
\label{sec:reasoning_impact}
\textit{How do reasoning-specialized models compare to general instruction-tuned models in Urdu?}

A central goal of UrduBench is to evaluate whether explicitly reasoning-oriented models outperform general instruction-tuned models in Urdu. As shown in Figure~\ref{fig:model_scaling_2}, reasoning-specialized models such as Phi-4-Mini-Reasoning and DeepSeek-R1-Distill variants exhibit clear gains on MGSM, particularly under Chain-of-Thought prompting, indicating that explicit reasoning supervision partially transfers to low-resource languages.
However, these advantages are less consistent on CommonSenseQA and OpenBookQA, where large instruction-tuned models such as Gemma-3-12B-it and Llama-3.1-8B-Instruct often outperform reasoning-specialized counterparts. This pattern suggests that reasoning supervision alone is insufficient without strong multilingual grounding, and that performance on commonsense and factual reasoning tasks in Urdu depends more on pretraining diversity and language alignment than on explicit reasoning objectives alone.

\subsection{Language Confusion Test on Reasoning Models}
\label{sec:language_impact}
\textit{How well do models maintain Urdu language consistency in their responses?}

\begin{table}[t]
\centering
\small
\renewcommand{\arraystretch}{1.4}
\caption{Language Confusion Test on MGSM Urdu CoT. Accuracy versus Answer language consistency for Urdu}
\setlength{\tabcolsep}{1pt}
\begin{tabular}{l|c|c}
\hline
\textbf{Model} & \textbf{Acc. (\%)} & \textbf{Cons. (\%)} \\
\hline
\hline
Gemma-3-4B-it & 64.0 & 98.8 \\

Falcon-h1-7b-instruct & 72.4 & 89.2 \\

Qwen2.5-7B-Instruct & 57.6 & 100.0 \\

Llama-3.1-8B-Instruct & 62.8 & 100.0 \\

Ministral-8B-Instruct-2410 & 58.4 & 98.8 \\

Gemma2-9B-it & 73.6 & 99.6 \\

Gemma-3-12B-it & 80.4 & 100.0 \\


DeepSeek-R1-Distill-Qwen-14B & 59.2 & 82.0 \\

\hline
\end{tabular}

\label{tab:language_confusion}
\footnotesize\begin{flushleft}
~~~~~~~~Higher consistency indicates lower language confusion or codeswitching.
\end{flushleft}
\end{table}

Language confusion test measures whether a model produces its final answers consistently in the target language (Urdu) without unintended code-switching. For each generated response, special tokens and mathematical notation are removed, and language identification is applied to the cleaned text. An answer is counted as language-consistent only if Urdu is the sole detected language; very short responses are excluded to avoid unreliable detection.

Formally, {Answer Language Consistency formula is taken from the study~\cite{wang2025polymathevaluatingmathematicalreasoning} as:
\[
\text{Lang. Consistency (\%)} =
\frac{\#\ \text{answers in Urdu}}{\#\ \text{total answers}} \times 100
\]

Language confusion directly impacts reasoning performance in low-resource languages such as Urdu. As shown in Table~\ref{tab:language_confusion}, models with higher answer-level language consistency achieve stronger MGSM-CoT accuracy, while residual code-switching correlates with degraded performance. This effect is most pronounced in multi-step reasoning, where inconsistent language use disrupts intermediate computations. Although larger models such as Gemma-3-12B-it and Llama-3.1-8B-Instruct generally exhibit stronger language control, scale alone is insufficient: reasoning-oriented models with aggressive distillation or quantization often show reduced language consistency despite comparable size. Overall, stable language alignment is a critical prerequisite for robust multilingual reasoning.

\section{Conclusion}
In this study, We introduced UrduBench, to best of our knowledge, the first standardized reasoning-focused benchmark suite for Urdu, comprising translations of MGSM, OpenBookQA, CommonSenseQA, and MATH-500. Our contextually ensembled, with human-in-the-loop framework leverages four diverse translation systems combined with LLM based feature integration and native human validation. Our context-aware translation strategy ensures linguistic coherence by processing questions and answer choices as unified units.

Through comprehensive evaluation of multiple state-of-the-art reasoning and non-reasoning models, we provide the first large-scale standardized comparison of LLM reasoning capabilities in Urdu across mathematical reasoning, commonsense knowledge, and factual understanding. Beyond advancing Urdu NLP, our methodology establishes a scalable framework for adapting standardized benchmarks to other low-resource languages. We also refer to this reasoning benchmark suite as "Miyar", which in Urdu denotes a standard or benchmark, reflecting a measure of reasoning quality in Urdu.







\section*{Acknowledgments}
ChatGPT~5.2 was used for grammatical refinement. The authors take full responsibility for the technical content of this paper.



\bibliography{custom}

\begin{thebibliography}{44}
\providecommand{\natexlab}[1]{#1}

\bibitem[{Adeeba et~al.(2025)Adeeba, Dillon, Sajjad, and Bhatt}]{arshad2024urb}
Farah Adeeba, Brian Dillon, Hassan Sajjad, and Rajesh Bhatt. 2025.
\newblock \href {https://arxiv.org/abs/2508.01006} {Urblimp: A benchmark for evaluating the linguistic competence of large language models in urdu}.
\newblock \emph{Preprint}, arXiv:2508.01006.

\bibitem[{Adelani et~al.(2021)Adelani, Abbott, Neubig, D{'}souza, Kreutzer, Lignos, Palen-Michel, Buzaaba, Rijhwani, Ruder, Mayhew, Azime, Muhammad, Emezue, Nakatumba-Nabende, Ogayo, Anuoluwapo, Gitau, Mbaye, Alabi, Yimam, Gwadabe, Ezeani, Niyongabo, Mukiibi, Otiende, Orife, David, Ngom, Adewumi, Rayson, Adeyemi, Muriuki, Anebi, Chukwuneke, Odu, Wairagala, Oyerinde, Siro, Bateesa, Oloyede, Wambui, Akinode, Nabagereka, Katusiime, Awokoya, MBOUP, Gebreyohannes, Tilaye, Nwaike, Wolde, Faye, Sibanda, Ahia, Dossou, Ogueji, DIOP, Diallo, Akinfaderin, Marengereke, and Osei}]{adelani2021masakhaner}
David~Ifeoluwa Adelani, Jade Abbott, Graham Neubig, Daniel D{'}souza, Julia Kreutzer, Constantine Lignos, Chester Palen-Michel, Happy Buzaaba, Shruti Rijhwani, Sebastian Ruder, Stephen Mayhew, Israel~Abebe Azime, Shamsuddeen~H. Muhammad, Chris~Chinenye Emezue, Joyce Nakatumba-Nabende, Perez Ogayo, Aremu Anuoluwapo, Catherine Gitau, Derguene Mbaye, and 42 others. 2021.
\newblock \href {https://doi.org/10.1162/tacl_a_00416} {{M}asakha{NER}: Named entity recognition for {A}frican languages}.
\newblock \emph{Transactions of the Association for Computational Linguistics}, 9:1116--1131.

\bibitem[{Ahmad et~al.(2025)Ahmad, Iqbal, Ahsan, Naeem, Khan, Riaz, Manzoor, Wang, and Nakov}]{ahmad-etal-2025-urdufactcheck}
Sarfraz Ahmad, Hasan Iqbal, Momina Ahsan, Numaan Naeem, Muhammad Ahsan~Riaz Khan, Arham Riaz, Muhammad~Arslan Manzoor, Yuxia Wang, and Preslav Nakov. 2025.
\newblock \href {https://doi.org/10.18653/v1/2025.findings-emnlp.1240} {{U}rdu{F}act{C}heck: An agentic fact-checking framework for {U}rdu with evidence boosting and benchmarking}.
\newblock In \emph{Findings of the Association for Computational Linguistics: EMNLP 2025}, pages 22788--22802, Suzhou, China. Association for Computational Linguistics.

\bibitem[{Arif et~al.(2024)Arif, Azeemi, Raza, and Athar}]{arif2024generalists}
Samee Arif, Abdul~Hameed Azeemi, Agha~Ali Raza, and Awais Athar. 2024.
\newblock \href {https://doi.org/10.18653/v1/2024.findings-emnlp.426} {Generalists vs. specialists: Evaluating large language models for {U}rdu}.
\newblock In \emph{Findings of the Association for Computational Linguistics: EMNLP 2024}, pages 7263--7280, Miami, Florida, USA. Association for Computational Linguistics.

\bibitem[{Basit et~al.(2024)Basit, Azeemi, and Raza}]{basit-etal-2024-challenges}
Abdul Basit, Abdul~Hameed Azeemi, and Agha~Ali Raza. 2024.
\newblock \href {https://doi.org/10.18653/v1/2024.loresmt-1.4} {Challenges in {U}rdu machine translation}.
\newblock In \emph{Proceedings of the Seventh Workshop on Technologies for Machine Translation of Low-Resource Languages (LoResMT 2024)}, pages 44--49, Bangkok, Thailand. Association for Computational Linguistics.

\bibitem[{Bhowmik et~al.(2025)Bhowmik, Dipto, Islam, Hsu, and Reasat}]{bhowmik2025evaluatingllmsmultilingualcapabilities}
Shimanto Bhowmik, Tawsif~Tashwar Dipto, Md~Sazzad Islam, Sheryl Hsu, and Tahsin Reasat. 2025.
\newblock \href {https://arxiv.org/abs/2507.23248} {Evaluating llms' multilingual capabilities for bengali: Benchmark creation and performance analysis}.
\newblock \emph{Preprint}, arXiv:2507.23248.

\bibitem[{Comanici et~al.(2025)Comanici, Bieber, Schaekermann, Pasupat, Sachdeva, Dhillon, Blistein, Ram, Zhang, Rosen, Marris, Petulla, Gaffney, Aharoni, Lintz, Pais, Jacobsson, Szpektor, Jiang, Haridasan, Omran, Saunshi, Bahri, Mishra, Chu, Boyd, Hekman, Parisi, Zhang, Kawintiranon, Bedrax-Weiss, Wang, Xu, Purkiss, Mendlovic, Deutel, Nguyen, Langley, Korn, Rossazza, Ramé, Waghmare, Miller, Byrd, Sheshan, Hadsell, Bhardwaj, Janus, Rissa, Horgan, Abdagic, Belenki, Allingham, Singh, Guidroz, Srinivasan, Schmit, Chiafullo, Elisseeff, Jha, Kolhar, Berrada, Ding, Si, Mallick, Och, Erell, Ni, Latkar, Yang, Sirkovic, Feng, Leland, Hornung, Wu, Blundell, Alvari, Huang, Yip, Deur, Liu, Surita, Duque, Damen, Jia, Guez, Mircea, Sinha, Magni, Stradomski, Marian, Galić, Chen, Husain, Singhal, Grewe, Aubet, Song, Blanco, Rechis, Ho, Munoz, Zheng, Hamrick, Mather, Taitelbaum, Rutherford, Lei, Chen, Shukla, Moreira, Doi, Isik, Shabat, Rogozińska, Kolipaka, Chang, Vušak, Venkatachary, Noghabi, Bharti, Jun, Zaks, Green,
  Challagundla, Wong, Mohammad, Hirsch, Cheng, Naim, Proleev, Vincent, Singh, Krikun, Krishnan, Ghahramani, Atias, Aggarwal, Kirov, Vytiniotis, Koh, Chronopoulou, Dogra, Ion, Tyen, Lee, Weissenberger, Strohman, Balakrishna, Rae, Velic, de~Liedekerke, Elyada, Yuan, Liu, Shani, Kishchenko, Alessio, Li, Song, Kwei, Jankowski, Pappu, Namiki, Ma, Tripuraneni, Cherry, Ikonomidis, Ling, Ji, Westberg, Wright, Yu, Parkinson, Ramaswamy, Connor, Yeganeh, Grover, Kenwright, Litchev, Apps, Tomala, Halim, Castro-Ros, Li, Boral, Sho, Yarom, Malmi, Klinghoffer, Lin, Ansell, S, Zhao, Zuo, Santoro, Cheng, Demmessie, Liu, Brichtova, Culp, Braun, Graur, Ng, Mehta, Phillips, Sundberg, Godbole, Liu, Katariya, Rim, Seyedhosseini, Ammirati, Valfridsson, Malihi, Knight, Toor, Lampe, Ittycheriah, Chiang, Yeung, Fréchette, Rao, Wang, Srivastava, Zhang, Rhodes, Brand, Weesner, Figotin, Gimeno, Fellinger, Marcenac, Leal, Marcus, Cotruta, Cabrera, Luo, Garrette, Axelrod, Baltateanu, Barker, Chen, Toma, Ingram, Riesa, Kulkarni, Zhang,
  Liu, Wang, Polacek, Wu, Hui, Reyes, Su, Barnes, Malhi, Siddiqui, Feng, Damaschin, Pighin, Steiner, Yang, Boppana, Ivanov, Kandoor, Shah, Mujika, Huang, Choquette-Choo, Patel, Yu, Creswell, Jerry, Liu, Barros, Razeghi, Roy, Culliton, Xiong, Pan, Strohmann, Powell, Seal, DeCarlo, Shyam, Katircioglu, Wang, Hardin, Odisho, Broder, Chang, Nair, Shtefan, O'Brien, Agarwal, Potluri, Goyal, Jhindal, Thakur, Stuken, Lyon, Toutanova, Feng, Wu, Horn, Wang, Cullum, Taubman, Shrivastava, Shi, Tomlinson, Patel, Tu, Oflazer, Pongetti, Yang, Taïga, Perot, Pierse, Han, Drori, Iturrate, Chakrabarti, Yeung, Dopson, ting Chen, Kulshreshtha, Guo, Pham, Schuster, Chen, Polozov, Xing, Zhou, Kacham, Kukliansky, Miech, Yaroshenko, Chi, Douglas, Fei, Blondel, Myla, Madmoni, Wu, Keysers, Kjems, Albuquerque, Yu, D'sa, Plantan, Ionescu, Elias, Gupta, Vuyyuru, Alcober, Zhou, Ji, Hartmann, Puttagunta, Song, Amid, Stefanoiu, Lee, Pucciarelli, Wang, Raul, Petrov, Tian, Anklin, Nti, Gomes, Schumacher, Vesom, Panagopoulos, Bousmalis, Andor,
  Jacob, Zhang, Rosgen, Kecman, Tung, Belias, Goodman, Covington, Wieder, Saxena, Davoodi, Huang, Maddineni, Roulet, Campbell-Ajala, Sessa, Xintian, Wu, Lai, Collins, Haig, Sakenas, Xu, Giustina, Shafey, Charoenpanit, Garg, Ainslie, Severson, Arenas, Pathak, Rajayogam, Feng, Bakker, Li, Wichers, Rogers, Geng, Li, Jagerman, Jia, Olmert, Sharon, Mauger, Mariserla, Ma, Mohabey, Kim, Andreev, Pollom, Love, Jain, Agrawal, Schroecker, Fortin, Warmuth, Liu, Leach, Blok, Girirajan, Aharoni, Uria, Sozanschi, Goldberg, Ionita, Ribeiro, Zlocha, Birodkar, Lachgar, Yuan, Choudhury, Ginsberg, Zheng, Dibb, Graves, Lokhande, Rasskin, Muraru, Quick, Tata, Sermanet, Chawla, Karo, Wang, Zhang, Keller, Dragan, Su, Chou, Liu, Tao, Prabhakara, Wilson, Liu, Wang, Evans, Du, Castaño, Prasad, Mahdy, Gerlach, Reid, Kahn, Zait, Pillai, Ulrich, Wang, Wassenberg, Farkash, Yalasangi, Wang, Bauza, Bucher, Liu, Yan, Leung, Sindhwani, Barnes, Singh, Jurin, Chang, Bhumihar, Eiger, Citovsky, Withbroe, Li, Xue, Santo, Stoyanov, Raimond, Zheng,
  Gao, Listík, Kwasiborski, Saputro, Ozturel, Mallya, Majmundar, West, Caron, Wei, Castrejon, Vikram, Ramachandran, Dhawan, Park, Smoot, van~den Driessche, Blau, Malik, Liang, Hirsch, dos Santos, Weinstein, van~den Oord, Lall, FitzGerald, Jiang, Yang, Webster, Elqursh, Pope, Rotival, Raposo, Zhu, Dean, Alabed, Tran, Gupta, Gleicher, Austin, Rosseel, Umekar, Das, Sun, Chen, Misiunas, Zhou, Di, Loo, Newlan, Li, Ramasesh, Xu, Chen, Gandhe, Soricut, Gupta, Hu, El-Sayed, Garcia, Brusilovsky, Chen, Bolt, Huang, Gurney, Zhang, Pritzel, Wilkiewicz, Seybold, Shamanna, Fischer, Dean, Gill, Mcilroy, Bhowmick, Selier, Yang, Cheng, Magay, Tan, Varma, Walder, Kocisky, Nakashima, Natsev, Kwong, Gog, Zhang, Dieleman, Jimma, Ryabtsev, Brahma, Steiner, Du, Žužul, Žanić, Raghavachari, Gierke, Zheng, Petrova, Dauphin, Liu, Kessler, Hand, Duvarney, Kim, Lee, Hussenot, Hui, Smith, Jain, Xia, Tomar, Amiri, Phan, Fuchs, Weyand, Tomasev, Cordell, Liu, Mallinson, Joshi, Crawford, Suggala, Chien, Fernando, Sanchez-Vargas,
  Williams, Crone, Luo, Karpov, Shan, Thurk, Strudel, Voigtlaender, Patil, Dozat, Khodaei, Singla, Ambroszczyk, Wu, Chang, Roark, Hegde, Ding, Filos, Wu, Pinto, Liu, Khanna, Pandey, Mcloughlin, Li, Haves, Zhou, Buchatskaya, Leal, de~Boursac, Akazawa, Anderson, Chen, Somandepalli, Liang, Goenka, Winkler, Grushetsky, Ding, Smith, Ye, Pont-Tuset, Li, Li, Golany, Wegner, Jiang, Barak, Shangguan, Vértes, Wong, Bornschein, Tudor, Bevilacqua, Schaul, Rawat, Zhao, Axiotis, Meng, McLean, Lai, Beattie, Kushman, Liu, Kutzman, Lang, Ye, Netrapalli, Mishra, Khan, Goel, Willoughby, Tian, Zhuang, Chen, Tsai, Kementsietsidis, Khare, Keeling, Xu, Waters, Altché, Popat, Mittal, Saxton, Badawy, Mathieu, Zheng, Zhou, Ranka, Shin, Duan, Salimans, Mihailescu, Shaham, Chang, Assael, Dikkala, Izzard, Cohen-Addad, Graves, Feinberg, Chung, Strouse, Karmon, Sharifzadeh, Ashwood, Pham, Blanton, Vasiloff, Barber, Geller, Zhou, Zubach, Huang, Zhang, Gupta, Young, Proskurnia, Votel, Gabeur, Barcik, Tripathi, Yu, Yan, Changpinyo,
  Pavetić, Coyle, Fujii, Mendez, Zhou, Rajamani, Hechtman, Cao, Juan, Tan, Dalibard, Du, Clay, Yao, Jia, Vijaykumar, Zhou, Bai, Hung, Pecht, Todorov, Khadke, Gupta, Lahoti, Autef, Duddu, Lee-Thorp, Bykovsky, Misiunas, Flennerhag, Thangaraj, McGiffin, Nado, Kunesch, Noever, Hertz, Liang, Stone, Palmer, Daruki, Pramanik, Põder, Kyker, Khan, Sluzhaev, Ritter, Ruderman, Zhou, Nagpal, Vodrahalli, Necula, Barham, Pavlick, Hartford, Shafran, Zhao, Mikuła, Eccles, Shimokawa, Garg, Vilnis, Chen, Shumailov, Lee, Abdelhamed, Xie, Cohen, Hlavnova, Malkin, Sitawarin, Lottes, Coquinot, Yu, Kumar, Zhang, Mahendru, Ahmed, Martens, Chen, Boag, Peng, Devin, Klimovskiy, Phuong, Vainstein, Xie, Ramabhadran, Howard, Yu, Goswami, Cui, Shleifer, Pinto, Yeh, Yang, Javanmardi, Ethier, Lee, Orbay, Kotecha, Bromberg, Shaw, Thornton, Rosenthal, Gu, Thomas, Gemp, Ayyar, Ushio, Selvan, Wee, Liu, Majzoubi, Yu, Abernethy, Liechty, Pan, Nguyen, Qiong, Hu, Perrin, Arora, Pitler, Wang, Shivakumar, Prost, Limonchik, Wang, Gao, Cour, Buch,
  Gui, Ivanova, Neubeck, Chan, Kim, Chen, Goyal, Chung, Liu, Su, Petrushkina, Shen, Joulin, Xu, Lin, Kulizhskaya, Chelba, Vasudevan, Collins, Bashlovkina, Lu, Fritz, Park, Zhou, Su, Tanburn, Sushkov, Rasquinha, Li, Prendki, Li, LV, Sharma, Fitoussi, Huang, Dai, Dao, Burrows, Prior, Qin, Pundak, Sjoesund, Khurshudov, Zhu, Webson, Kemp, Tan, Agrawal, Sargsyan, Cheng, Stephan, Kwiatkowski, Reid, Byravan, Michaely, Heess, Zhou, Goenka, Carpenter, Levskaya, Wang, Roberts, Leblond, Chikkerur, Ginzburg, Chang, Riachi, Chuqiao, Xu, Borsos, Pliskin, Pawar, Lustman, Kirkwood, Anand, Chaudhary, Kalb, Milan, Augenstein, Goldie, Prince, Raman, Sun, Xia, Cohen, Huo, Camp, Ellis, Zilka, Torres, Patel, Arora, Chan, Adler, Ayoub, Liang, Jamil, Jiang, Baumgartner, Sun, Karov, Akulov, Zheng, Cai, Fantacci, Rubin, Acha, Wang, D'Souza, Sathyanarayana, Dai, Rowe, Simanovsky, Goldman, Kuang, Pan, Rosenberg, Rojas-Esponda, Dutta, Zeng, Jurenka, Farquhar, Bansal, Iqbal, Roelofs, Joung, Beak, Ryu, Poplin, Wu, Alayrac, Buthpitiya,
  Ronneberger, Habtegebriel, Li, Cavallaro, Wei, Bensky, Denk, Ganapathy, Stanway, Joshi, Bertolini, Lo, Ma, Charles, Sampemane, Sahni, Chen, Askham, Gaddy, Young, Tan, Eyal, Bražinskas, Zhong, Wu, Epstein, Bailey, Hard, Lee, Goldshtein, Ruiz, Badawi, Lochbrunner, Kearns, Brown, Pardo, Weber, Yang, Jiang, Akin, Fu, Wainwright, Zou, Gaba, Manzagol, Kan, Song, Zainullina, Lin, Ko, Deshmukh, Jindal, Svensson, Tyam, Zhao, Kaeser-Chen, Baird, Moradi, Hall, Guo, Tsang, Liang, Pereira, Ganesh, Korotkov, Adamek, Thiagarajan, Tran, Chen, Tar, Jain, Dasgupta, Bilal, Reitter, Zhao, Vezzani, Gehman, Mehta, Beltrone, Dotiwalla, Guadarrama, Abbas, Karp, Georgiev, Ferng, Brockschmidt, Peng, Hirnschall, Verma, Bi, Xiao, Dabush, Xu, Wallis, Parker, Wang, Xu, Safarli, Tewari, Zhang, Kim, Gesmundo, Thomas, Levi, Chowdhury, Rao, Garst, Conway-Rahman, Ran, McKinney, Xiao, Yu, Agrawal, Stjerngren, Ionescu, Chen, Sharma, Chiu, Liu, Franko, Sanford, Cai, Michel, Ganapathy, Labanowski, Garrett, Vargas, Sun, Gale, Buschmann,
  Desjardins, Ghelani, Jain, Verma, Asawaroengchai, Eisenschlos, Harlalka, Kazawa, Metzler, Howland, Jian, Ades, Shah, Gangwani, Lee, Ring, Hernandez, Reich, Sinha, Sathe, Kovac, Gill, Kannan, D'olimpio, Sevenich, Whang, Kim, Sim, Chen, Zhang, Lall, Matias, Jia, Friesen, Nasso, Thapliyal, Perozzi, Yu, Shekhawat, Huda, Grabowski, Wang, Sreevatsa, Dib, Hassen, Schuh, Milutinovic, Welty, Quinn, Shah, Wang, Barth-Maron, Frye, Axelsson, Zhu, Ma, Giannoumis, Sedghi, Ye, Luan, Aydin, Chandra, Sampathkumar, Huang, Lavrenko, Eleryan, Hong, Hansen, Carthy, Samanta, Ćevid, Wang, Li, Voznesensky, Hoffman, Terzis, Sehwag, Fidel, He, Cai, He, Feng, Nikoltchev, Phatale, Chase, Lawton, Zhang, Ouyang, Tragut, Manshadi, Narayanan, Shen, Gao, Bolukbasi, Roy, Li, Golovin, Panait, Qin, Han, Anthony, Kudugunta, Patraucean, Ray, Chen, Yang, Bhatia, Talluri, Morris, Ražnatović, Brownfield, An, Peng, Kane, Zheng, Duduta, Kessinger, Noraky, Liu, Rong, Veličković, Rush, Goldin, Wei, Garlapati, Pantofaru, Kwon, Ni, Noland, Trapani,
  Beaufays, Roy, Chow, Turker, Cideron, Mei, Clark, Dou, Bošnjak, Leith, Du, Yazdanbakhsh, Nasr, Kwak, Sheth, Kaskasoli, Anand, Lakshminarayanan, Jerome, Bieber, Chu, Senges, Shen, Sridhar, Ndebele, Beyret, Mohamed, Chen, Freitag, Guo, Liu, Roit, Chen, Yan, Stone, Co-Reyes, Cole, Scellato, Azizi, Hashemi, Jin, Iyer, Valentine, György, Ahuja, Diaz, Lee, Clement, Kong, Garmon, Watts, Bhatia, Gupta, Miecnikowski, Vallet, Taly, Loper, Joshi, Atwood, Chick, Collier, Iliopoulos, Trostle, Gunel, Leal-Cavazos, Hrafnkelsson, Guzman, Ju, Forbes, Emond, Chauhan, Caine, Xiao, Zeng, Moufarek, Murphy, Meng, Gupta, Riedel, Das, Lawal, Narayan, Sosea, Swirhun, Friso, Neyshabur, Lu, Girgin, Wunder, Yvinec, Pyne, Carbune, Rijhwani, Guo, Doshi, Briukhov, Bain, Hitron, Wang, Gupta, Chen, Du, Zhang, Shah, Akula, Dylla, Kachra, Kuo, Zou, Wang, Xu, Zhu, Snyder, Menon, Firat, Mordatch, Yuan, Ponomareva, Blevins, Moore, Wang, Chen, Scholz, Dwornik, Lin, Li, Antognini, I, Song, Miller, Kalra, Raveret, Akerlund, Wu, Nystrom, Godbole,
  Liu, DeBalsi, Zhao, Liu, Caciularu, Lax, Khandelwal, Langston, Bailey, Lattanzi, Wang, Kovelamudi, Mondal, Guruganesh, Hua, Roval, Wesołowski, Ingale, Halcrow, Sohn, Angermueller, Raad, Stickgold, Lu, Kosik, Xie, Lillicrap, Huang, Zhang, Paulus, Farabet, Wertheim, Wang, Joshi, ling Ko, Wu, Agrawal, Lin, Sheng, Sung, Breland-King, Butterfield, Gawde, Singh, Zhang, Apte, Shetty, Hutter, Li, Salesky, Lebron, Kanerva, Paganini, Nguyen, Vallu, Peter, Velury, Kao, Hoover, Bortsova, Bishop, Jakobovits, Agostini, Agarwal, Liu, Kwong, Tavakkol, Bica, Greve, GP, Marcus, Hou, Duerig, Moroshko, Lacey, Davis, Amelot, Wang, Kim, Strinopoulos, Wan, Lan, Krishnan, Tang, Humphreys, Bai, Shtacher, Machado, Pang, Burke, Liu, Aravamudhan, Song, Hirst, Singh, Jou, Bai, Piccinno, Fu, Alazard, Meiri, Winter, Chen, Zhang, Heitkaemper, Lambert, Lee, Frömmgen, Rogulenko, Nair, Niemczyk, Bulyenov, Xu, Shemtov, Zadimoghaddam, Toropov, Wirth, Dai, Gollapudi, Zheng, Kurakin, Lee, Bullard, Serrano, Balazevic, Li, Schalkwyk, Murphy,
  Zhang, Sequeira, Datta, Agrawal, Sutton, Attaluri, Chiang, Farhan, Thornton, Lin, Choma, Nguyen, Dasgupta, Robinson, Comşa, Riley, Pillai, Mustafa, Golan, Zandieh, Lespiau, Porter, Ross, Rajayogam, Agarwal, Venugopalan, Shahriari, Yan, Xu, Tobin, Dubov, Shi, Recasens, Kovsharov, Borgeaud, Dery, Vasanth, Gribovskaya, Qiu, Mahdieh, Skut, Nielsen, Zheng, Yu, Bostock, Gupta, Archer, Rawles, Davies, Svyatkovskiy, Tsai, Halpern, Reisswig, Wydrowski, Chang, Puigcerver, Taege, Li, Schnider, Li, Dena, Xu, Telang, Shi, Zen, Kastner, Ko, Subramaniam, Kumar, Blois, Dai, Wieting, Lu, Zeldes, Xie, Hauth, Ţifrea, Li, El-Husseini, Abolafia, Zhou, Ding, Ghalebikesabi, Guía, Maksai, Ágoston Weisz, Arik, Sukhanov, Świetlik, Jia, Yu, Wang, Brand, Bloxwich, Kirmani, Chen, Go, Sprechmann, Kannen, Carin, Sandhu, Edkins, Nooteboom, Gupta, Maggiore, Azizi, Pritch, Yin, Gupta, Tarlow, Smith, Ivanov, Babaeizadeh, Goel, Kambala, Chu, Kastelic, Liu, Soltau, Stone, Agrawal, Kim, Soparkar, Tadepalli, Bunyan, Soh, Kannan, Kim, Chen,
  Halumi, Roy, Wang, Sercinoglu, Gibson, Bhatnagar, Sano, von Dincklage, Ren, Mitrevski, Olšák, She, Doersch, Jilei, Wang, Liu, Tan, Yakar, Warkentin, Ramirez, Lebsack, Dillon, Mathews, Cobley, Wu, Chen, Simon, Nath, Sainath, Bendebury, Julian, Mankalale, Ćurko, Zacchello, Brown, Sodhia, Howard, Caelles, Gupta, Evans, Bulanova, Katzen, Goldenberg, Tsitsulin, Stanton, Schillings, Kovalev, Fry, Shah, Lin, Upadhyay, Li, Radpour, Maggioni, Xiong, Haas, Brennan, Kamath, Savinov, Nagrani, Yacovone, Kappedal, Andriopoulos, Lao, Li, Rozhdestvenskiy, Hashimoto, Audibert, Austin, Rodriguez, Ruoss, Honke, Karkhanis, Xiong, Wei, Huang, Leng, Premachandran, Bileschi, Evangelopoulos, Mensink, Pavagadhi, Teplyashin, Chang, Xue, Tanzer, Goldman, Patel, Li, Wiesner, Zheng, Stewart-Binks, Han, Li, Luo, Lenc, Lučić, Xue, Mullins, Guseynov, Chang, Galatzer-Levy, Zhang, Bingham, Hu, Hartman, Ma, Griffith, Irpan, Radebaugh, Yue, Fan, Ungureanu, Sorokin, Teufel, Li, Anil, Paparas, Wang, Lin, Peng, Shum, Petrovic, Brady,
  Nguyen, Macherey, Li, Singh, Yenugula, Iinuma, Chen, Kopparapu, Stern, Dave, Thekkath, Perot, Kumar, Li, Xiao, Bilotti, Bateni, Noble, Lee, Vázquez-Reina, Salazar, Yang, Wang, Gruzewska, Rao, Raghuram, Xu, Ben-David, Mei, Dalmia, Zhang, Liu, Bansal, Pankov, Schwarcz, Burns, Chan, Sanghai, Liang, Liang, He, Stuart, Narayanan, Zhu, Frank, Fatemi, Sabne, Lang, Bhattacharya, Settle, Wang, McMahan, Tacchetti, Soares, Hadian, Cabi, Chung, Putikhin, Li, Chen, Tarango, Michalewski, Kazemi, Masoom, Sheftel, Shivanna, Vadali, Comanescu, Reid, Moore, Neelakantan, Sander, Herzig, Rosenberg, Dehghani, Choi, Fink, Hayes, Ge, Weng, Ho, Karro, Krishna, Thiet, Skerry-Ryan, Eppens, Andreetto, Sarma, Bonacina, Ayan, Nawhal, Shan, Dusenberry, Thakoor, Gubbi, Nguyen, Tsarfaty, Albanie, Mitrović, Gandhi, Chen, Epasto, Stephanov, Jin, Gehman, Amini, Weber, Behbahani, Xu, Allamanis, Chen, Ott, Sha, Jastrzebski, Qi, Greene, Wu, Toki, Vlasic, Shapiro, Kotikalapudi, Shen, Saeki, Xie, Cassirer, Bharadwaj, Kiyono, Bhojanapalli,
  Rosenfeld, Ritter, Mao, Oliveira, Egyed, Bandemer, Parisotto, Kinoshita, Pluto, Maniatis, Li, Guo, Ghiasi, Tarbouriech, Chatterjee, Jin, Katrina, Xu, Palomaki, Arnold, Sewak, Piccinini, Sharma, Albrecht, Purser-haskell, Vaswani, Chen, Wisniewski, Cao, Aslanides, Phu, Sieb, Agubuzu, Zheng, Sohn, Selvi, Andreassen, Subudhi, Eruvbetine, Woodman, Mery, Krause, Ren, Ma, Luo, Chen, Fan, Griffiths, Schuler, Li, Zhang, Sarr, Luo, Patana, Watson, Naboulsi, Collins, Sidhwani, Hoogeboom, Silver, Caveness, Zhao, Rodriguez, Deines, Bai, Griffin, Tagliasacchi, Xue, Babbula, Pang, Ding, Shen, Peake, Crocker, Raghvendra, Swisher, Han, Singh, Wu, Pchelin, Munkhdalai, Alon, Bacon, Robles, Bulian, Johnson, Powell, Ferreira, Li, Benzing, Velimirović, Soyer, Kong, Tony, Nguyên, Yang, Liu, van Amersfoort, Gillick, Sun, Rauschmayr, Zhang, Zhan, Zhou, Frolov, Yang, Vnukov, Rouillard, Li, Mandhane, Fallen, Venkataraman, Hu, Brennan, Lee, Chang, Sundermeyer, Pan, Ke, Tong, Fabrikant, Bono, Gu, Foley, Mao, Delakis, Bhaswar,
  Frostig, Li, Zipori, Hope, Kozlova, Mishra, Djolonga, Schiff, Merey, Briakou, Morgan, Wan, Hassidim, Skerry-Ryan, Sengupta, Jasarevic, Kallakuri, Kunkle, Brennan, Lieber, Mansoor, Walker, Zhang, Xie, Žužić, Chukwuka, Druinsky, Cho, Yao, Naeem, Butt, Kim, Jia, Jordan, Lelkes, Kurzeja, Wang, Zhao, Over, Chakladar, Prasetya, Jha, Ganapathy, Cong, Shroff, Saroufim, Miryoosefi, Hammad, Nasir, Xi, Gao, Maeng, Hora, Cheng, Haghani, Lewenberg, Lu, Matysiak, Raisinghani, Wang, Baugher, Sukthankar, Giang, Schultz, Fiedel, Chen, Lee, Dey, Zheng, Paul, Smith, Ly, Wang, Bansal, Perz, Ricco, Blank, Keshava, Sharma, Chow, Lad, Jalan, Osindero, Swanson, Scott, Ilić, Li, Jonnalagadda, Soudagar, Xiong, Batsaikhan, Jarrett, Kumar, Shah, Lawlor, Waters, Graham, May, Ramos, Lefdal, Cankara, Cano, O'Donoghue, Borovik, Liu, Grimstad, Alnahlawi, Tsihlas, Hudson, Grigorev, Jia, Huang, Igwe, Lebedev, Tang, Krivokon, Garcia, Tan, Jia, Stys, Vashishth, Liang, Venkatraman, Gu, Kementsietsidis, Zhu, Jung, Bai, Hosseini, Ahmed,
  Gupta, Yuan, Ashraf, Nigam, Vasudevan, Awasthi, Gilady, Mariet, Eskander, Li, Hu, Garrido, Schlattner, Zhang, Saxena, Dević, Muralidharan, Murthy, Zhou, Choi, Wongpanich, Wang, Shah, Xu, Huang, Spencer, Chen, Cohan, Wang, Tompson, Wu, Haroun, Li, Huergo, Yang, Yin, Wendt, Bendersky, Chaabouni, Snaider, Ferret, Jindal, Thompson, Xue, Bishop, Phal, Sharma, Sung, Radhakrishnan, Shomrat, Ingle, Vij, Gilmer, Istin, Sobell, Lu, Nottage, Sadigh, Willcock, Zhang, Xu, Brown, Lee, Wang, Zhu, Tay, Kim, Gutierrez, Sharma, Xian, Seo, Cui, Pochernina, Baetu, Jastrzębski, Ly, Elhawaty, Suh, Sezener, Wang, Yuen, Tucker, Cai, Yang, Wang, Muzio, Qian, Yoo, Lockhart, McKee, Guo, Mehrotra, Mendonça, Mehta, Ben, Tekur, Mu, Zhu, Krakovna, Lee, Maschinot, Cevey, Choe, Bai, Srinivasan, Gasaway, Young, Siegler, Holtmann-Rice, Piratla, Baumli, Yogev, Hofer, van Hasselt, Grant, Chervonyi, Silver, Hogue, Agarwal, Wang, Singh, Flynn, Lipschultz, David, Bellot, Yang, Le, Graziano, Olszewska, Hui, Maurya, Parotsidis, Chen, Oguntebi,
  Kelley, Baddepudi, Mauerer, Shaw, Siegman, Yang, Shetty, Roy, Song, Stokowiec, Burnell, Savant, Busa-Fekete, Miao, Ghosh, MacDermed, Lippe, Dektiarev, Behrman, Mentzer, Nguyen, Wei, Verma, Knutsen, Dasari, Yan, Mitrichev, Wang, Shejwalkar, Austin, Sunkara, Potti, Virin, Wright, Liu, Riva, Pot, Kochanski, Le, Balasubramaniam, Dhar, Liao, Bloniarz, Shukla, Cole, Lee, Zhang, Kafle, Vashishtha, Mahmoudieh, Chen, Hoffmann, Srinivasan, Lago, Shalom, Wang, Elabd, Sharma, Oh, Kothawade, Le, Monteiro, Yang, Alarakyia, Geirhos, Mincu, Garnes, Kobayashi, Mariooryad, Krasowiak, Zhixin, Lai, Mourad, Wang, Bu, Aharoni, Chen, Goyal, Zubov, Bapna, Dabir, Kothari, Lamerigts, Cao, Shar, Yew, Kulkarni, Mahaarachchi, Joshi, Zhu, Lichtarge, Zhou, Muckenhirn, Selo, Vinyals, Chen, Brohan, Mehta, Cogan, Wang, Geri, Ko, Chen, Viola, Shivam, Wang, Elish, Popa, Pereira, Liu, Koster, Kim, Zhang, Ebrahimi, Talukdar, Zheng, Poklukar, Mikhalap, Johnson, Vijayakumar, Omernick, Dibb, Dubey, Hu, Suman, Aggarwal, Kornakov, Xia, Lowe,
  Kolganov, Xiao, Nikolaev, Hemingray, Li, Iljazi, Rybiński, Sandhu, Lu, Luong, Jenatton, Govindaraj, Hui, Li, Dulac-Arnold, Park, Wang, Modi, Pouget-Abadie, Greller, Gupta, Berry, Ramachandran, Xie, McCafferty, Wang, Gupta, Lim, Bratanič, Brock, Akolzin, Sproch, Karliner, Kim, Goedeckemeyer, Shazeer, Schmid, Calandriello, Bhatia, Choromanski, Montgomery, Dua, Ramalho, King, Gao, Nguyen, Lindner, Pitta, Johnson, Salama, Ardila, Han, Farnese, Odoom, Wang, Ding, Rink, Smith, Lehri, Cohen, Vats, He, Gopavarapu, Paszke, Patel, Gansbeke, Loher, Castro, Voitovich, von Glehn, George, Niklaus, Eaton-Rosen, Rakićević, Jue, Perel, Zhang, Bahat, Pouget, Xing, Huot, Shenoy, Bos, Coriou, Richter, Noy, Wang, Ontanon, Qin, Makarchuk, Hassabis, Li, Sharma, Venkatesan, Kemaev, Daniel, Huang, Shah, Ponce, Warren, Chen, Faruqui, Wu, Andačić, Payrits, McDuff, Hume, Cao, Tessler, Wang, Wang, Rendulic, Agustsson, Johnson, Lando, Howard, Padmanabhan, Daswani, Banino, Kilgore, Heek, Ji, Caceres, Li, Kassner, Vlaskin, Liu,
  Grills, Hou, Sukkerd, Cheon, Shetty, Markeeva, Stanczyk, Iyer, Gong, Gao, Gopalakrishnan, Blyth, Reynolds, Bhoopchand, Bilenko, Gharibian, Zayats, Faust, Singh, Ma, Jiao, Vijayanarasimhan, Aroyo, Yadav, Chakera, Kakarla, Meshram, Gregor, Botea, Senter, Jia, Kovacs, Sharma, Baur, Kang, He, Zhuo, Kostelac, Laish, Peng, O'Bryan, Kasenberg, Rao, Leurent, Zhang, Stevens, Salazar, Zhang, Lobov, Walker, Porter, Redshaw, Ke, Rao, Lee, Lam, Moffitt, Kim, Qiao, Koo, Dadashi, Song, Sundararajan, Xu, Kawamoto, Zhong, Barbu, Reddy, Verzetti, Li, Papamakarios, Klimczak-Plucińska, Cassin, Kavukcuoglu, Swavely, Vaucher, Zhao, Hemsley, Tschannen, Ge, Menghani, Yu, Ha, He, Wu, Song, Sterneck, Zinke, Calian, Marsden, Ruiz, Hessel, Gueta, Lee, Farris, Gupta, Li, Saleh, Misra, Xiao, Mendolicchio, Buttimore, Krayvanova, Nayakanti, Wiethoff, Pande, Mirhoseini, Lao, Liu, Hua, Chen, Malkov, Kalashnikov, Gupta, Audhkhasi, Zhai, Kopalle, Jain, Ofek, Meyer, Baatarsukh, Strejček, Qian, Freedman, Figueira, Sokolik, Bachem, Lin,
  Kharrat, Hidey, Xu, Duan, Li, Ersoy, Everett, Cen, Santamaria-Fernandez, Taubenfeld, Mackinnon, Deng, Zablotskaia, Viswanadha, Goel, Yates, Deng, Choy, Chen, Sinha, Mossin, Wang, Szlam, Hao, Rubenstein, Toksoz-Exley, Aperghis, Zhong, Ahn, Isard, Lacombe, Luisier, Anastasiou, Kalley, Prabhu, Dunleavy, Bijwadia, Mao-Jones, Chen, Pasumarthi, Wood, Dostmohamed, Hurley, Simsa, Parrish, Pajarskas, Harvey, Skopek, Kochinski, Rey, Rieser, Zhou, Lee, Acharya, Li, Jiang, Zhang, Gipson, Mahintorabi, Gelmi, Khajehnouri, Yeh, Lee, Matthey, Baker, Pham, Fu, Pak, Gupta, Vasconcelos, Sadovsky, Walker, Hsiao, Zochbauer, Marzoca, Velan, Zeng, Baechler, Driess, Jain, Huang, Tao, Maggs, Levine, Schneider, Gemzer, Petit, Han, Fisher, Zelle, Biles, Ie, Fadeeva, Liu, Franco, Collister, Zhang, Wang, Zhao, Kieliger, Shuster, Zhu, Gong, Chan, Sun, Basu, Zimmermann, Hayes, Bapna, Snoek, Yang, Datta, Abdallah, Kilgour, Li, Mah, Jun, Rivière, Karmarkar, Spalink, Huang, Gonzalez, Tran, Nowak, Palowitch, Chadwick, Talius, Mehta, Sellam,
  Fränken, Nicosia, He, Kini, Amos, Basu, Jobe, Shaw, Xu, Evans, Ikeda, Yan, Jin, Wang, Yadav, Labzovsky, Sampath, Ma, Schumann, Siddhant, Shah, Youssef, Agarwal, Dabney, Tonioni, Ambar, Li, Guyon, Li, Soergel, Fang, Karadzhov, Udrescu, Trinh, Raunak, Noury, Guo, Gupta, Finkelstein, Petek, Liang, Billock, Sun, Wood, Song, Yu, Matejovicova, Cohen, Andra, D'Ambrosio, Deng, Nallatamby, Songhori, Dangovski, Lampinen, Botadra, Hillier, Cao, Baddi, Kuncoro, Yoshino, Bhagatwala, Ranzato, Schaeffer, Liu, Ye, Sarvana, Nham, Kuang, Gao, Baek, Mittal, Wahid, Gergely, Ni, Feldman, Muir, Lamblin, Macherey, Dyer, Kilpatrick, Campos, Bhutani, Fort, Ahmad, Severyn, Chatziprimou, Ferludin, Dimarco, Kusupati, Heyward, Bahir, Villela, Millican, Marcus, Bahargam, Unlu, Roth, Wei, Gopal, Ghoshal, Lee, Lin, Lees, Lee, Hosseini, Fan, Neel, Wu, Altun, Cai, Piqueras, Woodward, Bissacco, Haykal, Bordbar, Sundaram, Hodkinson, Toyama, Polovets, Myers, Sinha, Levinboim, Krishnakumar, Chhaparia, Sholokhova, Gundavarapu, Jawahar, Qureshi,
  Hu, Momchev, Rahtz, Wu, S, Dhamdhere, Guo, Gupta, Eslami, Schain, Blokzijl, Welling, Orr, Bolelli, Perez-Nieves, Sirotenko, Prasad, Kar, Pigem, Terzi, Weisz, Ghosh, Mavalankar, Madeka, Daugaard, Adam, Shah, Berman, Tran, Baker, Andrejczuk, Chole, Raboshchuk, Mirzazadeh, Kagohara, Wu, Schallhart, Orlando, Wang, Rrustemi, Xiong, Liu, Vezer, Ramsden, yiin Chang, Mudgal, Li, Vieillard, Hoshen, Ahmad, Slone, Hua, Potikha, Rossini, Stritar, Prakash, Wang, Dong, Nazari, Nehoran, Tekelioglu, Li, Badola, Funkhouser, Li, Yerram, Ganeshan, Formoso, Langner, Shi, Li, Yamamori, Panda, Saade, Scarpati, Breaux, Carey, Zhou, Hsieh, Bridgers, Butryna, Gupta, Tulsyan, Woo, Eltyshev, Grathwohl, Parks, Benjamin, Panigrahy, Dodhia, Freitas, Sauer, Song, Alet, Tolins, Paduraru, Zhou, Albert, Zhang, Shu, Bansal, Nguyen, Globerson, Xiao, Manyika, Hennigan, Rong, Matak, Bakalov, Sharma, Sinopalnikov, Pierson, Roller, Brown, Gao, Fukuzawa, Ghafouri, Vassigh, Barr, Wang, Korsun, Jayaram, Ren, Zaman, Khan, Lunts, Deutsch, Uthus, Katz,
  Samsikova, Khalifa, Sethi, Sun, Tang, Alon, Luo, Yu, Nayyar, Petrini, Truong, Hellendoorn, Chinaev, Alberti, Wang, Hu, Mirrokni, Balashankar, Aharon, Mehta, Iscen, Kready, Manning, Mohananey, Chen, Tripathi, Wu, Petrovski, Hwang, Baeuml, Chandrakaladharan, Liu, Coaguila, Chen, Ma, Tafti, Tatineni, Spitz, Ye, Vicol, Rosca, Puigdomènech, Yahav, Ghemawat, Lin, Kirk, Nabulsi, Brin, Bohnet, Caluwaerts, Veerubhotla, Zheng, Dai, Petrov, Xu, Mehran, Xu, Zintgraf, Choi, Hombaiah, Thoppilan, Reddi, Lew, Li, Webster, Sawhney, Lamprou, Shakeri, Lunayach, Chen, Bagri, Salcianu, Chen, Donchev, Magister, Nørly, Rodrigues, Izo, Noga, Zou, Köppe, Zhou, Lee, Long, Eisenbud, Chen, Schenck, To, Zhong, Taropa, Truong, Levy, Martins, Zhang, Semturs, Zhang, Yakubovich, Moreno, McConnaughey, Lu, Redmond, Weerts, Bitton, Refice, Lacasse, Conmy, Tallec, Odell, Forbes-Pollard, Socala, Hoech, Kohli, Walton, Wang, Sazanovich, Zhu, Kapishnikov, Galt, Denton, Murdoch, Sikora, Mohamed, Wei, First, McConnell, Cobo, Qin, Avrahami, Balle,
  Watanabe, Louis, Kraft, Ariafar, Gu, Rives, Yoon, Rusu, Cobon-Kerr, Hahn, Luo, Yuvein, Zhu, Ahuja, Benenson, Kaufman, Yu, Hightower, Zhang, Ni, Hendricks, Wang, Yona, Jain, Barrio, Bhupatiraju, Velusamy, Dafoe, Riedel, Thomas, Yuan, Bellaiche, Panthaplackel, Kloboves, Jauhari, Akbulut, Davchev, Gladchenko, Madras, Chuklin, Hill, Yuan, Madhavan, Leonhard, Scandinaro, Chen, Niu, Douillard, Damoc, Onoe, Pedregosa, Bertsch, Leichner, Pagadora, Malmaud, Ponda, Twigg, Duzhyi, Shen, Wang, Garg, Chen, Evci, Lee, Liu, Kojima, Yamaguchi, Rajendran, Piergiovanni, Rajendran, Fornoni, Ibagon, Ragan, Khan, Blitzer, Bunner, Sun, Kosakai, Lundberg, Elue, Guu, Park, Park, Narayanaswamy, Wu, Mudigonda, Cohn, Mu, Kumar, Graesser, Zhang, Killam, Zhuang, Giménez, Jishi, Ley-Wild, Zhai, Osawa, Cedillo, Liu, Upadhyay, Sieniek, Sharma, Paine, Angelova, Addepalli, Parada, Majumder, Lamp, Kumar, Deng, Myaskovsky, Sabolić, Dudek, York, de~Chaumont~Quitry, Nie, Cattle, Gunjan, Piot, Khawaja, Bang, Wang, Khodadadeh, R, Rawlani,
  Powell, Lee, Griesser, Oh, Magalhaes, Li, Tokumine, Vogel, Hsu, BC, Jindal, Cohen, Yang, Yuan, de~Cesare, Bruguier, Xu, Roy, Jacovi, Belov, Arya, Meadowlark, Cohen-Ganor, Ye, Morris-Suzuki, Banzal, Song, Ponnuramu, Zhang, Scrivener, Zaiem, Rochman, Han, Ghazi, Lee, Drath, Suo, Girgis, Shenoy, Nguyen, Eck, Gupta, Yan, Carreira, Gulati, Sang, Mirylenka, Cooney, Chou, Ling, Fan, Coleman, Tubone, Kumar, Baldridge, Hernandez-Campos, Lazaridou, Besley, Yona, Bulut, Wellens, Pierigiovanni, George, Green, Han, Tao, Clark, You, Abdolmaleki, Fu, Chen, Chaugule, Chandorkar, Rahman, Thompson, Koanantakool, Bernico, Ren, Vlasov, Vassilvitskii, Kula, Liang, Kim, Huang, Ye, Lepikhin, and Helmholz}]{comanici2025gemini25pushingfrontier}
Gheorghe Comanici, Eric Bieber, Mike Schaekermann, Ice Pasupat, Noveen Sachdeva, Inderjit Dhillon, Marcel Blistein, Ori Ram, Dan Zhang, Evan Rosen, Luke Marris, Sam Petulla, Colin Gaffney, Asaf Aharoni, Nathan Lintz, Tiago~Cardal Pais, Henrik Jacobsson, Idan Szpektor, Nan-Jiang Jiang, and 3416 others. 2025.
\newblock \href {https://arxiv.org/abs/2507.06261} {Gemini 2.5: Pushing the frontier with advanced reasoning, multimodality, long context, and next generation agentic capabilities}.
\newblock \emph{Preprint}, arXiv:2507.06261.

\bibitem[{Dang et~al.(2024)Dang, Singh, D'souza, Ahmadian, Salamanca, Smith, Peppin, Hong, Govindassamy, Zhao, Kublik, Amer, Aryabumi, Campos, Tan, Kocmi, Strub, Grinsztajn, Flet-Berliac, Locatelli, Lin, Talupuru, Venkitesh, Cairuz, Yang, Chung, Ko, Shi, Shukayev, Bae, Piktus, Castagné, Cruz-Salinas, Kim, Crawhall-Stein, Morisot, Roy, Blunsom, Zhang, Gomez, Frosst, Fadaee, Ermis, Üstün, and Hooker}]{dang2024ayaexpansecombiningresearch}
John Dang, Shivalika Singh, Daniel D'souza, Arash Ahmadian, Alejandro Salamanca, Madeline Smith, Aidan Peppin, Sungjin Hong, Manoj Govindassamy, Terrence Zhao, Sandra Kublik, Meor Amer, Viraat Aryabumi, Jon~Ander Campos, Yi-Chern Tan, Tom Kocmi, Florian Strub, Nathan Grinsztajn, Yannis Flet-Berliac, and 26 others. 2024.
\newblock \href {https://arxiv.org/abs/2412.04261} {Aya expanse: Combining research breakthroughs for a new multilingual frontier}.
\newblock \emph{Preprint}, arXiv:2412.04261.

\bibitem[{DeepSeek-AI(2025)}]{deepseekai2025deepseekr1incentivizingreasoningcapability}
DeepSeek-AI. 2025.
\newblock \href {https://arxiv.org/abs/2501.12948} {Deepseek-r1: Incentivizing reasoning capability in llms via reinforcement learning}.
\newblock \emph{Preprint}, arXiv:2501.12948.

\bibitem[{Gala et~al.(2023)Gala, Chitale, AK, Gumma, Doddapaneni, Kumar, Nawale, Sujatha, Puduppully, Raghavan, Kumar, Khapra, Dabre, and Kunchukuttan}]{gala2023indictrans2highqualityaccessiblemachine}
Jay Gala, Pranjal~A. Chitale, Raghavan AK, Varun Gumma, Sumanth Doddapaneni, Aswanth Kumar, Janki Nawale, Anupama Sujatha, Ratish Puduppully, Vivek Raghavan, Pratyush Kumar, Mitesh~M. Khapra, Raj Dabre, and Anoop Kunchukuttan. 2023.
\newblock \href {https://arxiv.org/abs/2305.16307} {Indictrans2: Towards high-quality and accessible machine translation models for all 22 scheduled indian languages}.
\newblock \emph{Preprint}, arXiv:2305.16307.

\bibitem[{Grattafiori et~al.(2024)Grattafiori, Dubey, Jauhri, Pandey, Kadian, Al-Dahle, Letman, Mathur, Schelten, Vaughan, Yang, Fan, Goyal, Hartshorn, Yang, Mitra, Sravankumar, Korenev, Hinsvark, Rao, Zhang, Rodriguez, Gregerson, Spataru, Roziere, Biron, Tang, Chern, Caucheteux, Nayak, Bi, Marra, McConnell, Keller, Touret, Wu, Wong, Ferrer, Nikolaidis, Allonsius, Song, Pintz, Livshits, Wyatt, Esiobu, Choudhary, Mahajan, Garcia-Olano, Perino, Hupkes, Lakomkin, AlBadawy, Lobanova, Dinan, Smith, Radenovic, Guzmán, Zhang, Synnaeve, Lee, Anderson, Thattai, Nail, Mialon, Pang, Cucurell, Nguyen, Korevaar, Xu, Touvron, Zarov, Ibarra, Kloumann, Misra, Evtimov, Zhang, Copet, Lee, Geffert, Vranes, Park, Mahadeokar, Shah, van~der Linde, Billock, Hong, Lee, Fu, Chi, Huang, Liu, Wang, Yu, Bitton, Spisak, Park, Rocca, Johnstun, Saxe, Jia, Alwala, Prasad, Upasani, Plawiak, Li, Heafield, Stone, El-Arini, Iyer, Malik, Chiu, Bhalla, Lakhotia, Rantala-Yeary, van~der Maaten, Chen, Tan, Jenkins, Martin, Madaan, Malo, Blecher,
  Landzaat, de~Oliveira, Muzzi, Pasupuleti, Singh, Paluri, Kardas, Tsimpoukelli, Oldham, Rita, Pavlova, Kambadur, Lewis, Si, Singh, Hassan, Goyal, Torabi, Bashlykov, Bogoychev, Chatterji, Zhang, Duchenne, Çelebi, Alrassy, Zhang, Li, Vasic, Weng, Bhargava, Dubal, Krishnan, Koura, Xu, He, Dong, Srinivasan, Ganapathy, Calderer, Cabral, Stojnic, Raileanu, Maheswari, Girdhar, Patel, Sauvestre, Polidoro, Sumbaly, Taylor, Silva, Hou, Wang, Hosseini, Chennabasappa, Singh, Bell, Kim, Edunov, Nie, Narang, Raparthy, Shen, Wan, Bhosale, Zhang, Vandenhende, Batra, Whitman, Sootla, Collot, Gururangan, Borodinsky, Herman, Fowler, Sheasha, Georgiou, Scialom, Speckbacher, Mihaylov, Xiao, Karn, Goswami, Gupta, Ramanathan, Kerkez, Gonguet, Do, Vogeti, Albiero, Petrovic, Chu, Xiong, Fu, Meers, Martinet, Wang, Wang, Tan, Xia, Xie, Jia, Wang, Goldschlag, Gaur, Babaei, Wen, Song, Zhang, Li, Mao, Coudert, Yan, Chen, Papakipos, Singh, Srivastava, Jain, Kelsey, Shajnfeld, Gangidi, Victoria, Goldstand, Menon, Sharma, Boesenberg,
  Baevski, Feinstein, Kallet, Sangani, Teo, Yunus, Lupu, Alvarado, Caples, Gu, Ho, Poulton, Ryan, Ramchandani, Dong, Franco, Goyal, Saraf, Chowdhury, Gabriel, Bharambe, Eisenman, Yazdan, James, Maurer, Leonhardi, Huang, Loyd, Paola, Paranjape, Liu, Wu, Ni, Hancock, Wasti, Spence, Stojkovic, Gamido, Montalvo, Parker, Burton, Mejia, Liu, Wang, Kim, Zhou, Hu, Chu, Cai, Tindal, Feichtenhofer, Gao, Civin, Beaty, Kreymer, Li, Adkins, Xu, Testuggine, David, Parikh, Liskovich, Foss, Wang, Le, Holland, Dowling, Jamil, Montgomery, Presani, Hahn, Wood, Le, Brinkman, Arcaute, Dunbar, Smothers, Sun, Kreuk, Tian, Kokkinos, Ozgenel, Caggioni, Kanayet, Seide, Florez, Schwarz, Badeer, Swee, Halpern, Herman, Sizov, Guangyi, Zhang, Lakshminarayanan, Inan, Shojanazeri, Zou, Wang, Zha, Habeeb, Rudolph, Suk, Aspegren, Goldman, Zhan, Damlaj, Molybog, Tufanov, Leontiadis, Veliche, Gat, Weissman, Geboski, Kohli, Lam, Asher, Gaya, Marcus, Tang, Chan, Zhen, Reizenstein, Teboul, Zhong, Jin, Yang, Cummings, Carvill, Shepard, McPhie,
  Torres, Ginsburg, Wang, Wu, U, Saxena, Khandelwal, Zand, Matosich, Veeraraghavan, Michelena, Li, Jagadeesh, Huang, Chawla, Huang, Chen, Garg, A, Silva, Bell, Zhang, Guo, Yu, Moshkovich, Wehrstedt, Khabsa, Avalani, Bhatt, Mankus, Hasson, Lennie, Reso, Groshev, Naumov, Lathi, Keneally, Liu, Seltzer, Valko, Restrepo, Patel, Vyatskov, Samvelyan, Clark, Macey, Wang, Hermoso, Metanat, Rastegari, Bansal, Santhanam, Parks, White, Bawa, Singhal, Egebo, Usunier, Mehta, Laptev, Dong, Cheng, Chernoguz, Hart, Salpekar, Kalinli, Kent, Parekh, Saab, Balaji, Rittner, Bontrager, Roux, Dollar, Zvyagina, Ratanchandani, Yuvraj, Liang, Alao, Rodriguez, Ayub, Murthy, Nayani, Mitra, Parthasarathy, Li, Hogan, Battey, Wang, Howes, Rinott, Mehta, Siby, Bondu, Datta, Chugh, Hunt, Dhillon, Sidorov, Pan, Mahajan, Verma, Yamamoto, Ramaswamy, Lindsay, Lindsay, Feng, Lin, Zha, Patil, Shankar, Zhang, Zhang, Wang, Agarwal, Sajuyigbe, Chintala, Max, Chen, Kehoe, Satterfield, Govindaprasad, Gupta, Deng, Cho, Virk, Subramanian, Choudhury,
  Goldman, Remez, Glaser, Best, Koehler, Robinson, Li, Zhang, Matthews, Chou, Shaked, Vontimitta, Ajayi, Montanez, Mohan, Kumar, Mangla, Ionescu, Poenaru, Mihailescu, Ivanov, Li, Wang, Jiang, Bouaziz, Constable, Tang, Wu, Wang, Wu, Gao, Kleinman, Chen, Hu, Jia, Qi, Li, Zhang, Zhang, Adi, Nam, Yu, Wang, Zhao, Hao, Qian, Li, He, Rait, DeVito, Rosnbrick, Wen, Yang, Zhao, and Ma}]{grattafiori2024llama3herdmodels}
Aaron Grattafiori, Abhimanyu Dubey, Abhinav Jauhri, Abhinav Pandey, Abhishek Kadian, Ahmad Al-Dahle, Aiesha Letman, Akhil Mathur, Alan Schelten, Alex Vaughan, Amy Yang, Angela Fan, Anirudh Goyal, Anthony Hartshorn, Aobo Yang, Archi Mitra, Archie Sravankumar, Artem Korenev, Arthur Hinsvark, and 542 others. 2024.
\newblock \href {https://arxiv.org/abs/2407.21783} {The llama 3 herd of models}.
\newblock \emph{Preprint}, arXiv:2407.21783.

\bibitem[{Hasanaath et~al.(2025)Hasanaath, Alansari, Ashraf, Chafik, Luqman, and Ezzini}]{arareasoner}
Ahmed~Abul Hasanaath, Aisha Alansari, Ahmed Ashraf, Salmane Chafik, Hamzah Luqman, and Saad Ezzini. 2025.
\newblock \href {https://doi.org/10.18653/v1/2025.findings-emnlp.1028} {{A}ra{R}easoner: Evaluating reasoning-based {LLM}s for {A}rabic {NLP}}.
\newblock In \emph{Findings of the Association for Computational Linguistics: EMNLP 2025}, pages 18898--18914, Suzhou, China. Association for Computational Linguistics.

\bibitem[{Hassan et~al.(2026)Hassan, Ahmed, and Awais}]{hassan2026qalblargeststateofthearturdu}
Muhammad~Taimoor Hassan, Jawad Ahmed, and Muhammad Awais. 2026.
\newblock \href {https://arxiv.org/abs/2601.08141} {Qalb: Largest state-of-the-art urdu large language model for 230m speakers with systematic continued pre-training}.
\newblock \emph{Preprint}, arXiv:2601.08141.

\bibitem[{Hendrycks et~al.(2021)Hendrycks, Burns, Kadavath, Arora, Basart, Tang, Song, and Steinhardt}]{math500}
Dan Hendrycks, Collin Burns, Saurav Kadavath, Akul Arora, Steven Basart, Eric Tang, Dawn Song, and Jacob Steinhardt. 2021.
\newblock Math-500: A subset of the math benchmark.
\newblock \url{https://huggingface.co/datasets/HuggingFaceH4/MATH-500}.
\newblock 500-problem subset of the MATH benchmark introduced in ``Measuring Mathematical Problem Solving With the MATH Dataset.''.

\bibitem[{Huber et~al.(2025)Huber, Chang, Wen, Fedorov, Elgamal, Huang, Suda, Sankar, Vogeti, Wang, Gladkov, Tai, Elogeel, Hefny, Chandra, Aly, Kumar, Krishnamoorthi, and Sagar}]{huber2025mobilellmprotechnicalreport}
Patrick Huber, Ernie Chang, Wei Wen, Igor Fedorov, Tarek Elgamal, Hanxian Huang, Naveen Suda, Chinnadhurai Sankar, Vish Vogeti, Yanghan Wang, Alex Gladkov, Kai~Sheng Tai, Abdelrahman Elogeel, Tarek Hefny, Vikas Chandra, Ahmed Aly, Anuj Kumar, Raghuraman Krishnamoorthi, and Adithya Sagar. 2025.
\newblock \href {https://arxiv.org/abs/2511.06719} {Mobilellm-pro technical report}.
\newblock \emph{Preprint}, arXiv:2511.06719.

\bibitem[{Jiang et~al.(2023)Jiang, Sablayrolles, Mensch, Bamford, Chaplot, de~las Casas, Bressand, Lengyel, Lample, Saulnier, Lavaud, Lachaux, Stock, Scao, Lavril, Wang, Lacroix, and Sayed}]{jiang2023mistral7b}
Albert~Q. Jiang, Alexandre Sablayrolles, Arthur Mensch, Chris Bamford, Devendra~Singh Chaplot, Diego de~las Casas, Florian Bressand, Gianna Lengyel, Guillaume Lample, Lucile Saulnier, Lélio~Renard Lavaud, Marie-Anne Lachaux, Pierre Stock, Teven~Le Scao, Thibaut Lavril, Thomas Wang, Timothée Lacroix, and William~El Sayed. 2023.
\newblock \href {https://arxiv.org/abs/2310.06825} {Mistral 7b}.
\newblock \emph{Preprint}, arXiv:2310.06825.

\bibitem[{Kazi et~al.(2025)Kazi, Rahim, and Khoja}]{kazi-etal-2025-crossing}
Samreen Kazi, Maria Rahim, and Shakeel~Ahmed Khoja. 2025.
\newblock \href {https://aclanthology.org/2025.indonlp-1.17/} {Crossing language boundaries: Evaluation of large language models on {U}rdu-{E}nglish question answering}.
\newblock In \emph{Proceedings of the First Workshop on Natural Language Processing for Indo-Aryan and Dravidian Languages}, pages 141--151, Abu Dhabi. Association for Computational Linguistics.

\bibitem[{Koishekenov et~al.(2023)Koishekenov, Berard, and Nikoulina}]{nllb}
Yeskendir Koishekenov, Alexandre Berard, and Vassilina Nikoulina. 2023.
\newblock \href {https://doi.org/10.18653/v1/2023.acl-long.198} {Memory-efficient {NLLB}-200: Language-specific expert pruning of a massively multilingual machine translation model}.
\newblock In \emph{Proceedings of the 61st Annual Meeting of the Association for Computational Linguistics (Volume 1: Long Papers)}, pages 3567--3585, Toronto, Canada. Association for Computational Linguistics.

\bibitem[{Lamsiyah et~al.(2025)Lamsiyah, Zeinalipour, El~amrany, Brust, Maggini, Bouvry, and Schommer}]{arabicsense}
Salima Lamsiyah, Kamyar Zeinalipour, Samir El~amrany, Matthias Brust, Marco Maggini, Pascal Bouvry, and Christoph Schommer. 2025.
\newblock \href {https://aclanthology.org/2025.wacl-1.1/} {{A}rabic{S}ense: A benchmark for evaluating commonsense reasoning in {A}rabic with large language models}.
\newblock In \emph{Proceedings of the 4th Workshop on Arabic Corpus Linguistics (WACL-4)}, pages 1--11, Abu Dhabi, UAE. Association for Computational Linguistics.

\bibitem[{Liu et~al.(2026)Liu, Khandelwal, Subramanian, Jouault, Rastogi, Sadé, Jeffares, Jiang, Cahill, Gavaudan, Sablayrolles, Héliou, You, Ehrenberg, Lo, Eliseev, Calvi, Sooriyarachchi, Bout, Rozière, Monicault, Lanfranchi, Barreau, Courtot, Grattarola, Dabert, de~las Casas, Chane-Sane, Ahmed, Berrada, Ecrepont, Guinet, Novikov, Kunsch, Lample, Martin, Gupta, Ludziejewski, Rute, Studnia, Amar, Delas, Roberts, Yadav, Chandu, Jain, Aitchison, Fainsin, Blier, Zhao, Martin, Saulnier, Gao, Buyl, Jennings, Pellat, Prins, Poirée, Guillaumin, Dinot, Futeral, Darrin, Augustin, Chiquier, Schimpf, Grinsztajn, Gupta, Raghuraman, Bousquet, Duchenne, Wang, von Platen, Jacob, Wambergue, Kurylowicz, Muddireddy, Chagniot, Stock, Agrawal, Torroba, Sauvestre, Soletskyi, Menneer, Vaze, Barry, Gandhi, Waghjale, Gandhi, Ghosh, Mishra, Aithal, Antoniak, Scao, Cachet, Sorg, Lavril, Saada, Chabal, Foubert, Robert, Wang, Lawson, Bewley, Bewley, Edwards, Jamil, Tomasini, Nemychnikova, Phung, Maladière, Richard, Bouaziz, Li,
  Marshall, Li, Yang, Ouahidi, Wang, Tang, and Ramzi}]{liu2026ministral3}
Alexander~H. Liu, Kartik Khandelwal, Sandeep Subramanian, Victor Jouault, Abhinav Rastogi, Adrien Sadé, Alan Jeffares, Albert Jiang, Alexandre Cahill, Alexandre Gavaudan, Alexandre Sablayrolles, Amélie Héliou, Amos You, Andy Ehrenberg, Andy Lo, Anton Eliseev, Antonia Calvi, Avinash Sooriyarachchi, Baptiste Bout, and 101 others. 2026.
\newblock \href {https://arxiv.org/abs/2601.08584} {Ministral 3}.
\newblock \emph{Preprint}, arXiv:2601.08584.

\bibitem[{López~Caro(2023)}]{lópezcaro2023}
Álvaro López~Caro. 2023.
\newblock \href {https://hdl.handle.net/2445/214303} {Machine translation evaluation metrics benchmarking: from traditional mt to llms}.

\bibitem[{Mihaylov et~al.(2018)Mihaylov, Clark, Khot, and Sabharwal}]{obqa}
Todor Mihaylov, Peter Clark, Tushar Khot, and Ashish Sabharwal. 2018.
\newblock \href {https://arxiv.org/abs/1809.02789} {Can a suit of armor conduct electricity? a new dataset for open book question answering}.
\newblock \emph{Preprint}, arXiv:1809.02789.

\bibitem[{Nahin et~al.(2025)Nahin, Nandi, Sarker, Muhtaseem, Kowsher, Shill, Ibrahim, Menon, Muntasir, and Alam}]{nahin2025titullmsfamilybanglallms}
Shahriar~Kabir Nahin, Rabindra~Nath Nandi, Sagor Sarker, Quazi~Sarwar Muhtaseem, Md~Kowsher, Apu~Chandraw Shill, Md~Ibrahim, Mehadi~Hasan Menon, Tareq~Al Muntasir, and Firoj Alam. 2025.
\newblock \href {https://doi.org/10.18653/v1/2025.findings-acl.1279} {{T}itu{LLM}s: A family of {B}angla {LLM}s with comprehensive benchmarking}.
\newblock In \emph{Findings of the Association for Computational Linguistics: ACL 2025}, pages 24922--24940, Vienna, Austria. Association for Computational Linguistics.

\bibitem[{Olmo et~al.(2025)Olmo, Ettinger, Bertsch, Kuehl, Graham, Heineman, Groeneveld, Brahman, Timbers, Ivison, Morrison, Poznanski, Lo, Soldaini, Jordan, Chen, Noukhovitch, Lambert, Walsh, Dasigi, Berry, Malik, Shah, Geng, Arora, Gupta, Anderson, Xiao, Murray, Romero, Graf, Asai, Bhagia, Wettig, Liu, Rangapur, Anastasiades, Huang, Schwenk, Trivedi, Magnusson, Lochner, Liu, Miranda, Sap, Morgan, Schmitz, Guerquin, Wilson, Huff, Bras, Xin, Shao, Skjonsberg, Shen, Li, Wilde, Pyatkin, Merrill, Chang, Gu, Zeng, Sabharwal, Zettlemoyer, Koh, Farhadi, Smith, and Hajishirzi}]{olmo2025olmo3}
Team Olmo, Allyson Ettinger, Amanda Bertsch, Bailey Kuehl, David Graham, David Heineman, Dirk Groeneveld, Faeze Brahman, Finbarr Timbers, Hamish Ivison, Jacob Morrison, Jake Poznanski, Kyle Lo, Luca Soldaini, Matt Jordan, Mayee Chen, Michael Noukhovitch, Nathan Lambert, Pete Walsh, and 49 others. 2025.
\newblock \href {https://arxiv.org/abs/2512.13961} {Olmo 3}.
\newblock \emph{Preprint}, arXiv:2512.13961.

\bibitem[{OpenAI(2025)}]{openai}
OpenAI. 2025.
\newblock Openai.
\newblock \url{https://openai.com/}.
\newblock Accessed: 2025-12-04.

\bibitem[{Ortiz et~al.(2025)Ortiz, Melero, del Pozo, Conde, Reviriego, Rocher, and Grandury}]{11232191}
Irene~Plaza Ortiz, Nina Melero, Cristina del Pozo, Javier Conde, Pedro Reviriego, Marina~Mayor Rocher, and María Grandury. 2025.
\newblock \href {https://doi.org/10.1109/GACLM67198.2025.11232191} {Spanish and llm benchmarks: Is mmlu lost in translation?}
\newblock In \emph{2025 2nd International Generative AI and Computational Language Modelling Conference (GACLM)}, pages 104--108.

\bibitem[{Plaza et~al.(2024)Plaza, Melero, del Pozo, Conde, Reviriego, Mayor-Rocher, and Grandury}]{plaza2024spanishllmbenchmarksmmlu}
Irene Plaza, Nina Melero, Cristina del Pozo, Javier Conde, Pedro Reviriego, Marina Mayor-Rocher, and María Grandury. 2024.
\newblock \href {https://arxiv.org/abs/2406.17789} {Spanish and llm benchmarks: is mmlu lost in translation?}
\newblock \emph{Preprint}, arXiv:2406.17789.

\bibitem[{Qwen et~al.(2025)Qwen, :, Yang, Yang, Zhang, Hui, Zheng, Yu, Li, Liu, Huang, Wei, Lin, Yang, Tu, Zhang, Yang, Yang, Zhou, Lin, Dang, Lu, Bao, Yang, Yu, Li, Xue, Zhang, Zhu, Men, Lin, Li, Tang, Xia, Ren, Ren, Fan, Su, Zhang, Wan, Liu, Cui, Zhang, and Qiu}]{qwen2025qwen25technicalreport}
Qwen, :, An~Yang, Baosong Yang, Beichen Zhang, Binyuan Hui, Bo~Zheng, Bowen Yu, Chengyuan Li, Dayiheng Liu, Fei Huang, Haoran Wei, Huan Lin, Jian Yang, Jianhong Tu, Jianwei Zhang, Jianxin Yang, Jiaxi Yang, Jingren Zhou, and 25 others. 2025.
\newblock \href {https://arxiv.org/abs/2412.15115} {Qwen2.5 technical report}.
\newblock \emph{Preprint}, arXiv:2412.15115.

\bibitem[{Shafique et~al.(2025)Shafique, Mehreen, Arham, Amjad, Butt, and Farooq}]{alif8b}
Muhammad~Ali Shafique, Kanwal Mehreen, Muhammad Arham, Maaz Amjad, Sabur Butt, and Hamza Farooq. 2025.
\newblock \href {https://arxiv.org/abs/2510.09051} {Alif: Advancing urdu large language models via multilingual synthetic data distillation}.
\newblock \emph{Preprint}, arXiv:2510.09051.

\bibitem[{Shi et~al.(2023)Shi, Suzgun, Freitag, Wang, Srivats, Vosoughi, Chung, Tay, Ruder, Zhou, Das, and Wei}]{mgsm}
Freda Shi, Mirac Suzgun, Markus Freitag, Xuezhi Wang, Suraj Srivats, Soroush Vosoughi, Hyung~Won Chung, Yi~Tay, Sebastian Ruder, Denny Zhou, Dipanjan Das, and Jason Wei. 2023.
\newblock \href {https://openreview.net/forum?id=fR3wGCk-IXp} {Language models are multilingual chain-of-thought reasoners}.
\newblock In \emph{The Eleventh International Conference on Learning Representations}.

\bibitem[{Song et~al.(2025)Song, Li, Lothritz, Ezzini, Sleem, Gentile, State, Bissyandé, and Klein}]{song2025smalllanguagemodelsilver}
Yewei Song, Lujun Li, Cedric Lothritz, Saad Ezzini, Lama Sleem, Niccolo Gentile, Radu State, Tegawendé~F. Bissyandé, and Jacques Klein. 2025.
\newblock \href {https://arxiv.org/abs/2503.24102} {Is small language model the silver bullet to low-resource languages machine translation?}
\newblock \emph{Preprint}, arXiv:2503.24102.

\bibitem[{Talmor et~al.(2019)Talmor, Herzig, Lourie, and Berant}]{csqa}
Alon Talmor, Jonathan Herzig, Nicholas Lourie, and Jonathan Berant. 2019.
\newblock \href {https://doi.org/10.18653/v1/N19-1421} {{C}ommonsense{QA}: A question answering challenge targeting commonsense knowledge}.
\newblock In \emph{Proceedings of the 2019 Conference of the North {A}merican Chapter of the Association for Computational Linguistics: Human Language Technologies, Volume 1 (Long and Short Papers)}, pages 4149--4158, Minneapolis, Minnesota. Association for Computational Linguistics.

\bibitem[{Team et~al.(2025)Team, Kamath, Ferret, Pathak, Vieillard, Merhej, Perrin, Matejovicova, Ramé, Rivière, Rouillard, Mesnard, Cideron, bastien Grill, Ramos, Yvinec, Casbon, Pot, Penchev, Liu, Visin, Kenealy, Beyer, Zhai, Tsitsulin, Busa-Fekete, Feng, Sachdeva, Coleman, Gao, Mustafa, Barr, Parisotto, Tian, Eyal, Cherry, Peter, Sinopalnikov, Bhupatiraju, Agarwal, Kazemi, Malkin, Kumar, Vilar, Brusilovsky, Luo, Steiner, Friesen, Sharma, Sharma, Gilady, Goedeckemeyer, Saade, Feng, Kolesnikov, Bendebury, Abdagic, Vadi, György, Pinto, Das, Bapna, Miech, Yang, Paterson, Shenoy, Chakrabarti, Piot, Wu, Shahriari, Petrini, Chen, Lan, Choquette-Choo, Carey, Brick, Deutsch, Eisenbud, Cattle, Cheng, Paparas, Sreepathihalli, Reid, Tran, Zelle, Noland, Huizenga, Kharitonov, Liu, Amirkhanyan, Cameron, Hashemi, Klimczak-Plucińska, Singh, Mehta, Lehri, Hazimeh, Ballantyne, Szpektor, Nardini, Pouget-Abadie, Chan, Stanton, Wieting, Lai, Orbay, Fernandez, Newlan, yeong Ji, Singh, Black, Yu, Hui, Vodrahalli, Greff, Qiu,
  Valentine, Coelho, Ritter, Hoffman, Watson, Chaturvedi, Moynihan, Ma, Babar, Noy, Byrd, Roy, Momchev, Chauhan, Sachdeva, Bunyan, Botarda, Caron, Rubenstein, Culliton, Schmid, Sessa, Xu, Stanczyk, Tafti, Shivanna, Wu, Pan, Rokni, Willoughby, Vallu, Mullins, Jerome, Smoot, Girgin, Iqbal, Reddy, Sheth, Põder, Bhatnagar, Panyam, Eiger, Zhang, Liu, Yacovone, Liechty, Kalra, Evci, Misra, Roseberry, Feinberg, Kolesnikov, Han, Kwon, Chen, Chow, Zhu, Wei, Egyed, Cotruta, Giang, Kirk, Rao, Black, Babar, Lo, Moreira, Martins, Sanseviero, Gonzalez, Gleicher, Warkentin, Mirrokni, Senter, Collins, Barral, Ghahramani, Hadsell, Matias, Sculley, Petrov, Fiedel, Shazeer, Vinyals, Dean, Hassabis, Kavukcuoglu, Farabet, Buchatskaya, Alayrac, Anil, Dmitry, Lepikhin, Borgeaud, Bachem, Joulin, Andreev, Hardin, Dadashi, and Hussenot}]{gemmateam2025gemma3technicalreport}
Gemma Team, Aishwarya Kamath, Johan Ferret, Shreya Pathak, Nino Vieillard, Ramona Merhej, Sarah Perrin, Tatiana Matejovicova, Alexandre Ramé, Morgane Rivière, Louis Rouillard, Thomas Mesnard, Geoffrey Cideron, Jean bastien Grill, Sabela Ramos, Edouard Yvinec, Michelle Casbon, Etienne Pot, Ivo Penchev, and 197 others. 2025.
\newblock \href {https://arxiv.org/abs/2503.19786} {Gemma 3 technical report}.
\newblock \emph{Preprint}, arXiv:2503.19786.

\bibitem[{Team et~al.(2024)Team, Riviere, Pathak, Sessa, Hardin, Bhupatiraju, Hussenot, Mesnard, Shahriari, Ramé, Ferret, Liu, Tafti, Friesen, Casbon, Ramos, Kumar, Lan, Jerome, Tsitsulin, Vieillard, Stanczyk, Girgin, Momchev, Hoffman, Thakoor, Grill, Neyshabur, Bachem, Walton, Severyn, Parrish, Ahmad, Hutchison, Abdagic, Carl, Shen, Brock, Coenen, Laforge, Paterson, Bastian, Piot, Wu, Royal, Chen, Kumar, Perry, Welty, Choquette-Choo, Sinopalnikov, Weinberger, Vijaykumar, Rogozińska, Herbison, Bandy, Wang, Noland, Moreira, Senter, Eltyshev, Visin, Rasskin, Wei, Cameron, Martins, Hashemi, Klimczak-Plucińska, Batra, Dhand, Nardini, Mein, Zhou, Svensson, Stanway, Chan, Zhou, Carrasqueira, Iljazi, Becker, Fernandez, van Amersfoort, Gordon, Lipschultz, Newlan, yeong Ji, Mohamed, Badola, Black, Millican, McDonell, Nguyen, Sodhia, Greene, Sjoesund, Usui, Sifre, Heuermann, Lago, McNealus, Soares, Kilpatrick, Dixon, Martins, Reid, Singh, Iverson, Görner, Velloso, Wirth, Davidow, Miller, Rahtz, Watson, Risdal,
  Kazemi, Moynihan, Zhang, Kahng, Park, Rahman, Khatwani, Dao, Bardoliwalla, Devanathan, Dumai, Chauhan, Wahltinez, Botarda, Barnes, Barham, Michel, Jin, Georgiev, Culliton, Kuppala, Comanescu, Merhej, Jana, Rokni, Agarwal, Mullins, Saadat, Carthy, Cogan, Perrin, Arnold, Krause, Dai, Garg, Sheth, Ronstrom, Chan, Jordan, Yu, Eccles, Hennigan, Kocisky, Doshi, Jain, Yadav, Meshram, Dharmadhikari, Barkley, Wei, Ye, Han, Kwon, Xu, Shen, Gong, Wei, Cotruta, Kirk, Rao, Giang, Peran, Warkentin, Collins, Barral, Ghahramani, Hadsell, Sculley, Banks, Dragan, Petrov, Vinyals, Dean, Hassabis, Kavukcuoglu, Farabet, Buchatskaya, Borgeaud, Fiedel, Joulin, Kenealy, Dadashi, and Andreev}]{gemmateam2024gemma2improvingopen}
Gemma Team, Morgane Riviere, Shreya Pathak, Pier~Giuseppe Sessa, Cassidy Hardin, Surya Bhupatiraju, Léonard Hussenot, Thomas Mesnard, Bobak Shahriari, Alexandre Ramé, Johan Ferret, Peter Liu, Pouya Tafti, Abe Friesen, Michelle Casbon, Sabela Ramos, Ravin Kumar, Charline~Le Lan, Sammy Jerome, and 179 others. 2024.
\newblock \href {https://arxiv.org/abs/2408.00118} {Gemma 2: Improving open language models at a practical size}.
\newblock \emph{Preprint}, arXiv:2408.00118.

\bibitem[{Thellmann et~al.(2024)Thellmann, Stadler, Fromm, Buschhoff, Jude, Barth, Leveling, Flores-Herr, Köhler, Jäkel, and Ali}]{thellmann2024multilingualllmevaluationeuropean}
Klaudia Thellmann, Bernhard Stadler, Michael Fromm, Jasper~Schulze Buschhoff, Alex Jude, Fabio Barth, Johannes Leveling, Nicolas Flores-Herr, Joachim Köhler, René Jäkel, and Mehdi Ali. 2024.
\newblock \href {https://arxiv.org/abs/2410.08928} {Towards multilingual llm evaluation for european languages}.
\newblock \emph{Preprint}, arXiv:2410.08928.

\bibitem[{Vaswani et~al.(2025)Vaswani, Callahan, Chaluvaraju, Gordić, Gupta, Jain, Mansingka, Monk, Nguyen, Parmar, Pust, Romanski, Rushton, Shehper, Shivaprasad, Singla, Smith, Srivastava, Thomas, Tripathy, Vanjani, Velingker, and {{Essential AI}}}]{rnj1_instruct}
Ashish Vaswani, Mike Callahan, Adarsh Chaluvaraju, Aleksa Gordić, Devaansh Gupta, Yash Jain, Divya Mansingka, Philip Monk, Khoi Nguyen, Mohit Parmar, Michael Pust, Tim Romanski, Peter Rushton, Ali Shehper, Divya Shivaprasad, Somanshu Singla, Kurt Smith, Saurabh Srivastava, Anil Thomas, and 4 others. 2025.
\newblock \href {https://huggingface.co/EssentialAI/rnj-1-instruct} {{Rnj-1-Instruct}}.
\newblock Instruction-tuned model release.

\bibitem[{Wang et~al.(2025)Wang, Zhang, Tang, Wei, Yang, Wang, Sun, Sun, Zhang, Wu, Cang, Zhang, Huang, Lin, Huang, and Zhou}]{wang2025polymathevaluatingmathematicalreasoning}
Yiming Wang, Pei Zhang, Jialong Tang, Haoran Wei, Baosong Yang, Rui Wang, Chenshu Sun, Feitong Sun, Jiran Zhang, Junxuan Wu, Qiqian Cang, Yichang Zhang, Fei Huang, Junyang Lin, Fei Huang, and Jingren Zhou. 2025.
\newblock \href {https://arxiv.org/abs/2504.18428} {Polymath: Evaluating mathematical reasoning in multilingual contexts}.
\newblock \emph{Preprint}, arXiv:2504.18428.

\bibitem[{Weerawardhena et~al.(2025)Weerawardhena, Kassianik, Nelson, Saglam, Vellore, Priyanshu, Vijay, Aufiero, Goldblatt, Burch, Li, He, Kedia, Oshiba, Yang, Singer, and Karbasi}]{weerawardhena2025llama31foundationaisecurityllm8binstructtechnicalreport}
Sajana Weerawardhena, Paul Kassianik, Blaine Nelson, Baturay Saglam, Anu Vellore, Aman Priyanshu, Supriti Vijay, Massimo Aufiero, Arthur Goldblatt, Fraser Burch, Ed~Li, Jianliang He, Dhruv Kedia, Kojin Oshiba, Zhouran Yang, Yaron Singer, and Amin Karbasi. 2025.
\newblock \href {https://arxiv.org/abs/2508.01059} {Llama-3.1-foundationai-securityllm-8b-instruct technical report}.
\newblock \emph{Preprint}, arXiv:2508.01059.

\bibitem[{Wu et~al.(2025)Wu, Shen, Su, and Xiong}]{cbench}
Junru Wu, Tianhao Shen, Linxi Su, and Deyi Xiong. 2025.
\newblock \href {https://doi.org/10.18653/v1/2025.findings-acl.1083} {{C}{\texttwosuperior}{RB}ench: A {C}hinese complex reasoning benchmark for large language models}.
\newblock In \emph{Findings of the Association for Computational Linguistics: ACL 2025}, pages 21031--21050, Vienna, Austria. Association for Computational Linguistics.

\bibitem[{Xiaomi et~al.(2025)Xiaomi, :, Xia, Shen, Cici, Zhu, Zhang, Wang, Zhang, Liu, Xiao, Dong, Zhao, Li, Wang, Yu, Chen, Wang, Ma, Deng, Huang, Song, Jiang, Ye, Cai, He, Zhang, Zhang, Wang, Tian, Zhao, Qu, Xu, Shi, Bao, Fang, Zhou, Zhou, Li, Zhu, Chen, Wang, Liu, Li, Gu, Ren, Liu, Deng, Zhuang, Lv, Yang, Zhang, Yong, Zhang, Song, Xu, Wang, Yan, Tu, Tian, Wang, Yu, Lin, Song, and Yue}]{xiaomi2025mimounlockingreasoningpotential}
LLM-Core Xiaomi, :, Bingquan Xia, Bowen Shen, Cici, Dawei Zhu, Di~Zhang, Gang Wang, Hailin Zhang, Huaqiu Liu, Jiebao Xiao, Jinhao Dong, Liang Zhao, Peidian Li, Peng Wang, Shihua Yu, Shimao Chen, Weikun Wang, Wenhan Ma, and 46 others. 2025.
\newblock \href {https://arxiv.org/abs/2505.07608} {Mimo: Unlocking the reasoning potential of language model -- from pretraining to posttraining}.
\newblock \emph{Preprint}, arXiv:2505.07608.

\bibitem[{Xu et~al.(2025)Xu, Peng, Awadalla, Chen, Chen, Gao, Kim, Li, Ren, Shen, Wang, Xu, Gao, and Chen}]{xu2025phi4minireasoningexploringlimitssmall}
Haoran Xu, Baolin Peng, Hany Awadalla, Dongdong Chen, Yen-Chun Chen, Mei Gao, Young~Jin Kim, Yunsheng Li, Liliang Ren, Yelong Shen, Shuohang Wang, Weijian Xu, Jianfeng Gao, and Weizhu Chen. 2025.
\newblock \href {https://arxiv.org/abs/2504.21233} {Phi-4-mini-reasoning: Exploring the limits of small reasoning language models in math}.
\newblock \emph{Preprint}, arXiv:2504.21233.

\bibitem[{Yang et~al.(2025)Yang, Li, Yang, Zhang, Hui, Zheng, Yu, Gao, Huang, Lv, Zheng, Liu, Zhou, Huang, Hu, Ge, Wei, Lin, Tang, Yang, Tu, Zhang, Yang, Yang, Zhou, Zhou, Lin, Dang, Bao, Yang, Yu, Deng, Li, Xue, Li, Zhang, Wang, Zhu, Men, Gao, Liu, Luo, Li, Tang, Yin, Ren, Wang, Zhang, Ren, Fan, Su, Zhang, Zhang, Wan, Liu, Wang, Cui, Zhang, Zhou, and Qiu}]{yang2025qwen3technicalreport}
An~Yang, Anfeng Li, Baosong Yang, Beichen Zhang, Binyuan Hui, Bo~Zheng, Bowen Yu, Chang Gao, Chengen Huang, Chenxu Lv, Chujie Zheng, Dayiheng Liu, Fan Zhou, Fei Huang, Feng Hu, Hao Ge, Haoran Wei, Huan Lin, Jialong Tang, and 41 others. 2025.
\newblock \href {https://arxiv.org/abs/2505.09388} {Qwen3 technical report}.
\newblock \emph{Preprint}, arXiv:2505.09388.

\bibitem[{Zhu et~al.(2024)Zhu, Liu, Dong, Xu, Huang, Kong, Chen, and Li}]{zhu-etal-2024-multilingual}
Wenhao Zhu, Hongyi Liu, Qingxiu Dong, Jingjing Xu, Shujian Huang, Lingpeng Kong, Jiajun Chen, and Lei Li. 2024.
\newblock \href {https://doi.org/10.18653/v1/2024.findings-naacl.176} {Multilingual machine translation with large language models: Empirical results and analysis}.
\newblock In \emph{Findings of the Association for Computational Linguistics: NAACL 2024}, pages 2765--2781, Mexico City, Mexico. Association for Computational Linguistics.

\bibitem[{Zuo et~al.(2025)Zuo, Velikanov, Chahed, Belkada, Rhayem, Kunsch, Hacid, Yous, Farhat, Khadraoui, Farooq, Campesan, Cojocaru, Djilali, Hu, Chaabane, Khanna, Seddik, Huynh, Khac, AlQadi, Mokeddem, Chami, Abubaker, Lubinets, Piskorski, and Frikha}]{zuo2025falconh1familyhybridheadlanguage}
Jingwei Zuo, Maksim Velikanov, Ilyas Chahed, Younes Belkada, Dhia~Eddine Rhayem, Guillaume Kunsch, Hakim Hacid, Hamza Yous, Brahim Farhat, Ibrahim Khadraoui, Mugariya Farooq, Giulia Campesan, Ruxandra Cojocaru, Yasser Djilali, Shi Hu, Iheb Chaabane, Puneesh Khanna, Mohamed El~Amine Seddik, Ngoc~Dung Huynh, and 8 others. 2025.
\newblock \href {https://arxiv.org/abs/2507.22448} {Falcon-h1: A family of hybrid-head language models redefining efficiency and performance}.
\newblock \emph{Preprint}, arXiv:2507.22448.

\end{thebibliography}

\newpage

\clearpage        
\onecolumn        

\appendix

\section{Prompts and Results Used in the Translation Framework}

This appendix provides the exact prompts used in our
Contextual Ensemble Translation framework with human-in-the-loop.
All translations in the benchmark were generated using the following
LLM prompts.

\subsection{Qwen 3 and Gemini 2.5 Pro Urdu Contextual Translation Prompt}
\label{a1}
The prompt below was used to generate Urdu translations directly
from English source texts for MGSM.

\begin{quote}
{\ttfamily Translate given text into Urdu. DO NOT respond the text.
ONLY output the translation of the text. Maintain natural,
fluent, and cultural context in Urdu translation.\par}
\end{quote}
For  OpenBookQA, and CommonSenseQA, following prompt was used using Context-aware strategy:
\begin{quote}
{\ttfamily You are a translation expert. Analyze the given translations
against the original text and create one improved translation
by combining the best features from all of them.\par}
\end{quote}
\begin{quote}
\ttfamily
Translate the following English text into Urdu. Return \textbf{only} a valid JSON object with two keys:

1. \texttt{"question"}: the translated question\\
\quad 2. \texttt{"choices"}: an array of translated choice texts (maintain the same order)

Maintain natural, fluent, and culturally appropriate Urdu. Do \textbf{not} include any explanation or extra text.

\textbf{Question:} \{question\_stem\}\\
\quad \textbf{Choices:}\\
\{choices\_formatted\}

\medskip
\textbf{Return format:}\\
\texttt{\{"question": "translated question", "choices": ["choice1", "choice2", ...]\}}
\end{quote}

For MATH-500, following prompt was used:
\begin{quote}
{\ttfamily You are an expert translator specializing in translating mathematical content from English to Urdu while preserving all LaTeX mathematical notation.

Task: Translate the following from English to Urdu.

Critical Rules:\\
1. NEVER translate or modify ANY LaTeX math expressions — keep all content inside \$ or \$\$ EXACTLY unchanged\\
2. NEVER translate mathematical symbols, variables, or formulas\\
3. Only translate natural language text\\
4. Preserve exact structure, line breaks, and spacing\\
5. Use fluent, formal academic Urdu\\
6. Preserve punctuation and special characters\\

\par}
\end{quote}

\subsection{GPT-5.1 Prompt for Best Feature Selection}
\label{a2}
GPT-5.1 was used to merge, refine, and improve candidate translations
generated by IndicTrans2, Qwen-3-30B, Gemini 2.5 Pro, and NLLB.

\begin{quote}
{\ttfamily You are a translation expert. Analyze the given translations
against the original text and create one improved translation
by combining the best features from all of them.\par}
\end{quote}

\subsection{Translation and Heuristic Processing Algorithm}
\label{a3}
The following algorithm describes the translation and heuristic processing pipeline applied to all translated samples as described in Algorithm~\ref{alg:urdu-quality}.

\subsection{Contextually Ensembled Translation
Method Stats}
\label{a4}

\begin{table*}[h!]
\centering
\small
\renewcommand{\arraystretch}{1.4}
\caption{Human Validation Win--Loss Statistics Across Translation Candidates}
\setlength{\tabcolsep}{8pt}
\begin{tabular}{l|cccc|c}
\hline
\textbf{Dataset} & \textbf{IndicTrans2} & \textbf{NLLB-200} & \textbf{Qwen-3-30B} & \textbf{Gemini-2.5-Pro} & \textbf{Total} \\
\hline
MGSM & 3 & 2 & 1 & \textbf{125} & 131 \\
CommonSenseQA & 74 & 52 & 17 & \textbf{826} & 969 \\
OpenBookQA & 36 & 27 & 9 & \textbf{313} & 385 \\
MATH-500 & 3 & 2 & 51 & \textbf{312} & 368 \\
\hline
\end{tabular}
\label{tab:human_win_loss}
\footnotesize
\begin{flushleft}
Each cell reports the number of instances for which a translation candidate was selected as the final output by native Urdu-speaking annotators. Higher counts indicate stronger human preference.
\end{flushleft}
\end{table*}


\begin{figure}[t]
\centering
\begin{tikzpicture}
\begin{axis}[
    xbar stacked,
    width=0.85\textwidth,
    height=0.5\textwidth,
    xlabel={Number of Wins},
    xlabel style={font=\small},
    ylabel={},
    ylabel style={font=\small},
    xmin=0,
    xmax=260,
    ytick=data,
    yticklabels={MGSM, CommonSenseQA, OpenBookQA, MATH-500},
    tick label style={font=\small},
    legend style={
        at={(0.5,-0.2)},
        anchor=north,
        legend columns=2,
        font=\small,
        /tikz/every even column/.append style={column sep=0.5cm}
    },
    bar width=15pt,
    enlarge y limits=0.2,
    nodes near coords,
    nodes near coords style={font=\footnotesize},
    grid=major,
    grid style={line width=.1pt, draw=gray!30},
    major grid style={line width=.2pt, draw=gray!50},
]

\addplot[fill=blue!50, draw=blue!80!black] coordinates {
    (97,0) (205,1) (86,2) (67,3)
};

\addplot[fill=orange!70, draw=orange!80!black] coordinates {
    (22,0) (47,1) (29,2) (65,3)
};

\legend{GPT-5.1 Refined Wins, Human-Edited Wins}

\end{axis}


\end{tikzpicture}
\caption{Human preference win--loss comparison between GPT-5.1 refined and human-edited Urdu translations across four datasets. The stacked bars show the number of wins for each translation method, with win rate percentages displayed on the right.}
\label{fig:human_win_loss}
\end{figure}
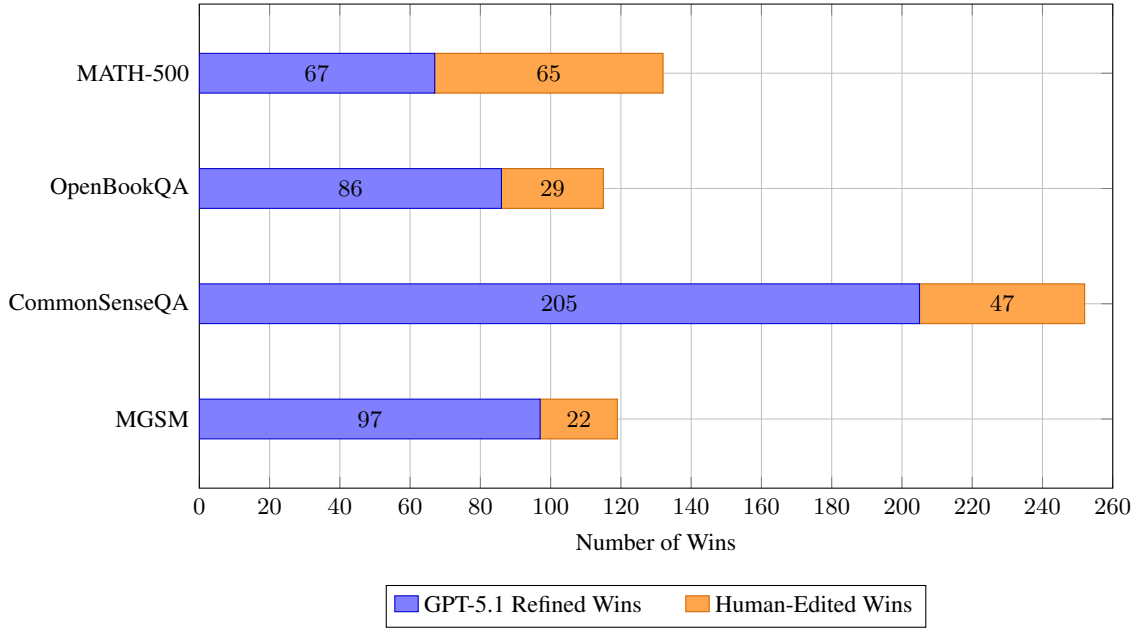


\begin{algorithm}[htbp]
\caption{Multi-Model Translation with Heuristic Validation and Consensus}
\label{alg:urdu-quality}
\footnotesize

\begin{algorithmic}[1]

\REQUIRE 
Dataset $D$,\\
translation models $\mathcal{M}$=\{\text{IndicTrans2}, \text{NLLB}, \text{Gemini-2.5-Pro}, \text{Qwen-3}\},\\
maximum retries $K=5$,\\ 
minimum length $L_{min} = 10$,\\
english content ratio threshold $\tau = 0.3$
\ENSURE For each sample $x$, translations
$\mathcal{T}_x = \{t^{(1)}, t^{(2)}, t^{(3)}, t^{(4)}, t^{GPT}\}$

\FOR{each sample $x \in D$}

    \STATE $T \leftarrow \emptyset$ \COMMENT{Validated model translations}

    \FOR{each model $m \in \mathcal{M}$ \textbf{in parallel}}
        \STATE $attempt \leftarrow 0$
        \STATE $valid \leftarrow false$

        \WHILE{$attempt < K$ \AND $valid = false$}
            \STATE $x_m \leftarrow \textsc{Translate}(x, m,$ \text{prompt}$)$

            \IF{required fields missing in $x_m$}
                \STATE Record missing field error
            \ENDIF

            \STATE $q^{ur} \leftarrow$ Urdu question in $x_m$
            \IF{$q^{ur}$ is empty or $|q^{ur}| < L_{min}$}
                \STATE Record invalid Urdu question
            \ENDIF

            \IF{no Urdu characters in $q^{ur}$ or English ratio $> \tau$}
                \STATE Record mixed-language question
            \ENDIF

            \STATE $C^{ur} \leftarrow$ Urdu answer choices in $x_m$
            \IF{$C^{ur}$ missing or number of choices incorrect}
                \STATE Record invalid Urdu choices
            \ENDIF

            \FOR{each choice $c \in C^{ur}$}
                \IF{no Urdu characters in $c$ or English ratio $> \tau$}
                    \STATE Record invalid Urdu choice
                \ENDIF
            \ENDFOR

            \IF{answer key missing or invalid}
                \STATE Record answer key error
            \ENDIF

            \IF{no errors recorded}
                \STATE $T \leftarrow T \cup \{x_m\}$
                \STATE $valid \leftarrow true$
            \ELSE
                \STATE $attempt \leftarrow attempt + 1$
            \ENDIF
        \ENDWHILE

        \IF{$valid = false$}
            \STATE $T \leftarrow T \cup \{x_m\}$
        \ENDIF
    \ENDFOR

    \STATE $t^{GPT} \leftarrow \textsc{GPT-5.1}(D, T,$ prompt$)$

    \STATE $\mathcal{T}_x \leftarrow T \cup \{t^{GPT}\}$
    \STATE Save($\mathcal{T}_x$)

\ENDFOR

\end{algorithmic}
\end{algorithm}

\clearpage        

\section{Context-Aware Translation Examples}
\label{appendix:context-transaltion}
This appendix provides illustrative examples demonstrating how context-aware translation improves the grammatical coherence and semantic alignment of Urdu multiple-choice questions. Table~\ref{tab:at1} compares translations generated with and without contextual information, highlighting the benefits of translating the full question–choice sequence as a unified input.

\begin{table*}[h!]
    \centering
    \includegraphics[width=\textwidth]{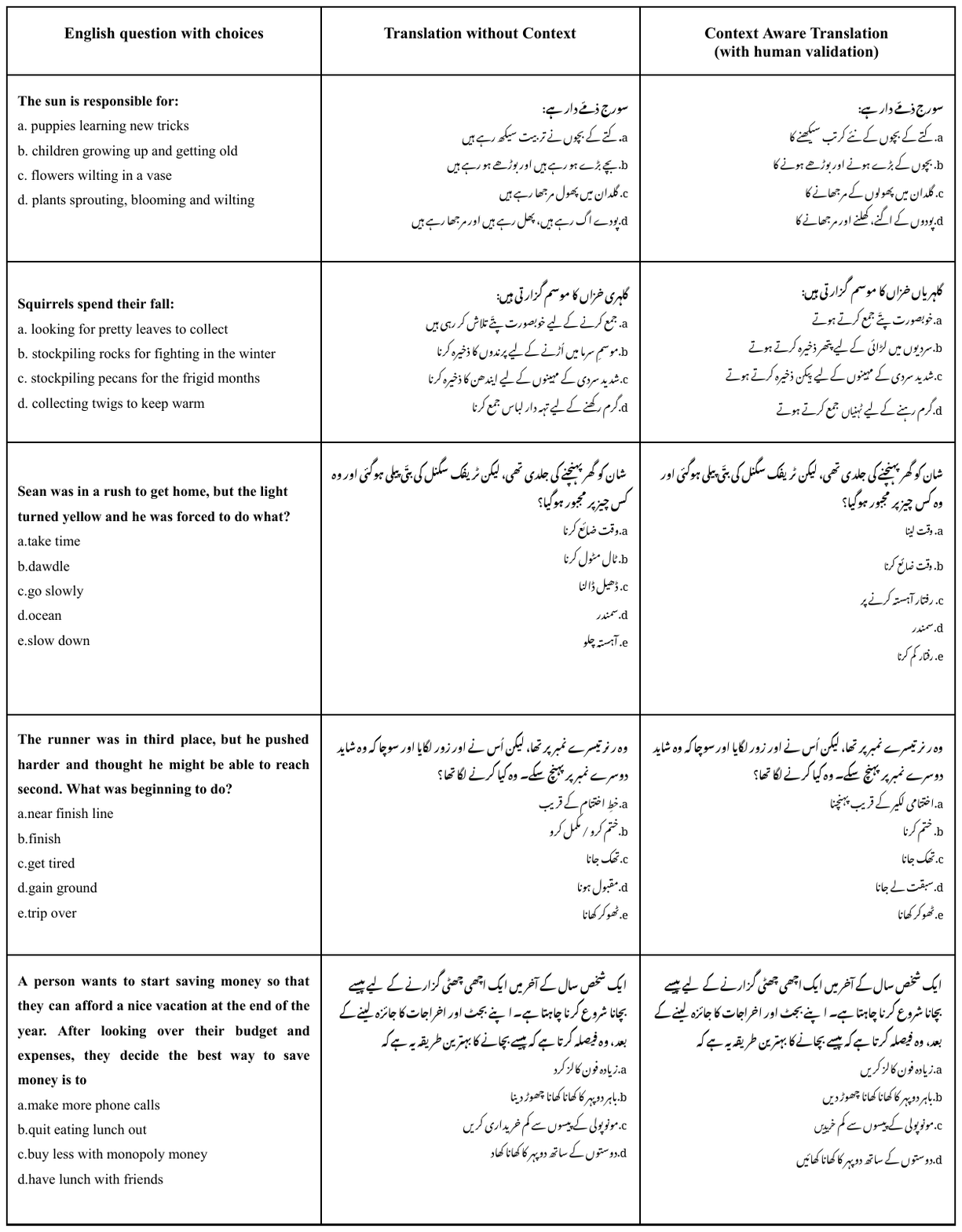}
    \caption{Examples of with and without Context Translation}
    \label{tab:at1}
\end{table*}

\clearpage        

\section{Human Annotation Guidelines}
\label{appendix:human-guidlines}
The rubrics provided in Table \ref{tab:refinement-guidelines}  were given to native Urdu-speaking annotators to guide their evaluation and selection of the best translation candidate from the ensemble outputs.

\begin{table*}[h!]
    \centering
    \includegraphics[width=\textwidth]{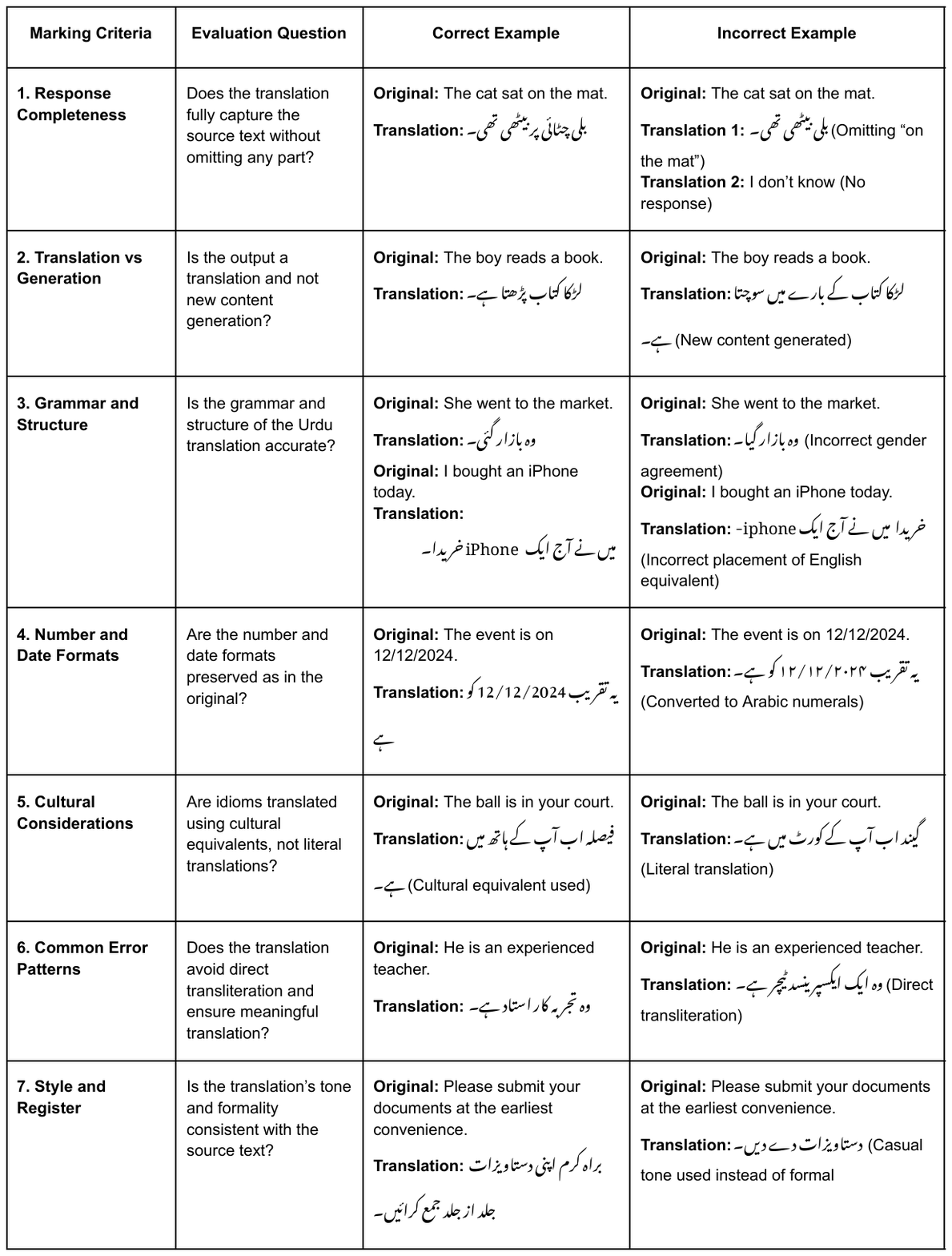}
    \caption{Overview of the Urdu-translated datasets refinement guidelines.}
    \label{tab:refinement-guidelines}
\end{table*}

\clearpage        




\clearpage

\section{Evaluation Prompts and Configurations}
\label{app:prompts}

All evaluation prompts are fixed across models and prompting strategies, including Direct, Chain-of-Thought (CoT), and Few-Shot + CoT (FS+CoT), to ensure comparability. Evaluations are conducted using \texttt{lm-eval-harness} with a \texttt{vLLM} backend and tensor parallelism set to 4. 

\begin{figure*}[h!]
    \centering
    \includegraphics[width=\textwidth]{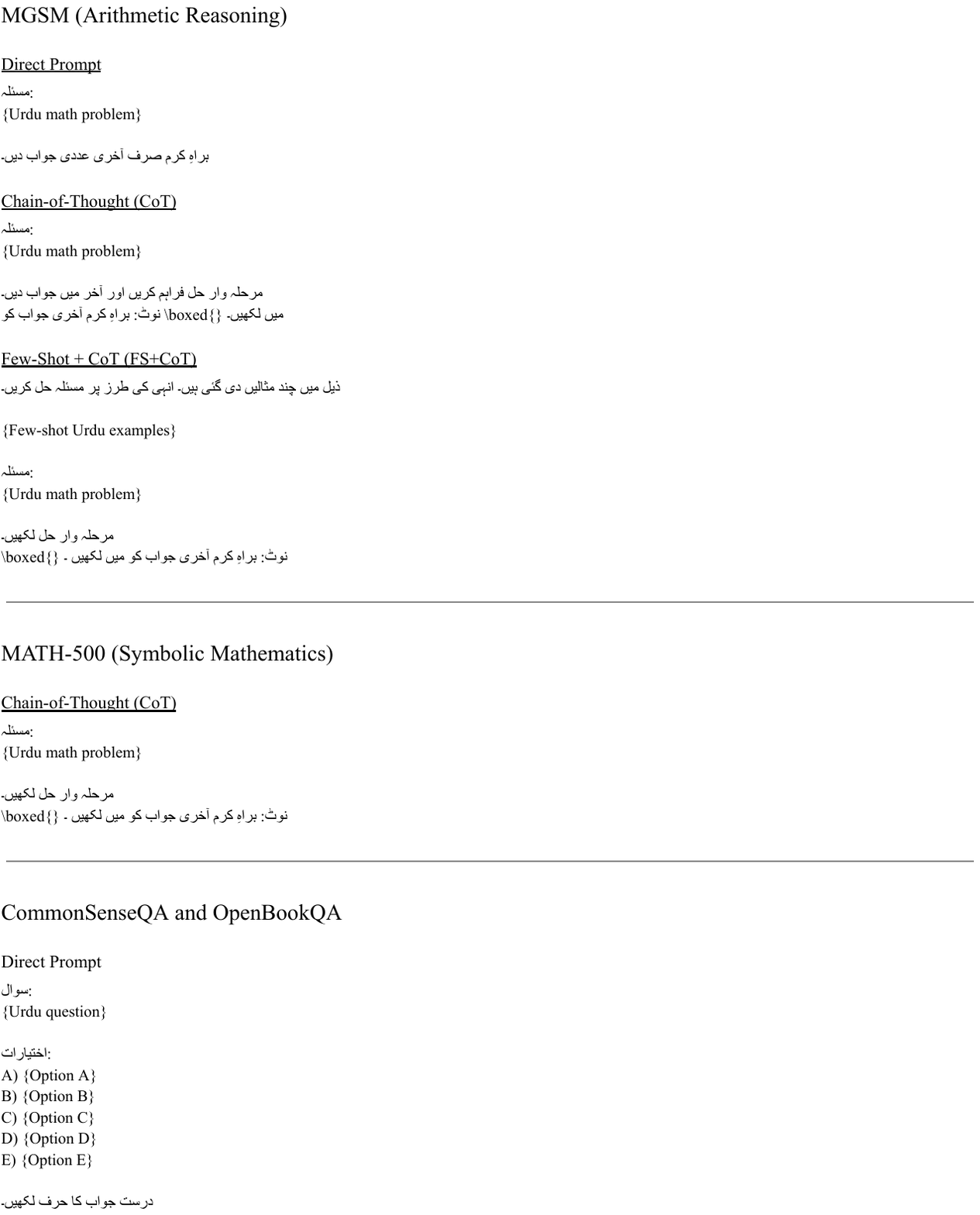}
    \caption{Evaluation prompt for Urdu MGSM, MATH-500, CommonSenseQA, and OpenBookQA}
    \label{fig:refinement-guidelines}
\end{figure*}





\begin{table}[t]
\centering
\small
\renewcommand{\arraystretch}{1.3}
\begin{tabularx}{\linewidth}{l X X X}
\toprule
\textbf{$<$7B} & \textbf{7B} & \textbf{8B} & \textbf{$>$8B} \\
\midrule
MobileLLM-Pro (1B) 
& Falcon-h1-7B-Instruct 
& Alif-1.0-8B-Instruct  
& Gemma2-9B-it \\

\cite{huber2025mobilellmprotechnicalreport}& \cite{zuo2025falconh1familyhybridheadlanguage} &\cite{alif8b} & \cite{gemmateam2024gemma2improvingopen} \\

LLaMA-3.2-3B-Instruct 
& MiMo-7B-Base 
& Aya-Expanse-8B 
& Gemma-3-12B-it \\

\cite{grattafiori2024llama3herdmodels}& \cite{xiaomi2025mimounlockingreasoningpotential}& \cite{dang2024ayaexpansecombiningresearch}& \cite{gemmateam2025gemma3technicalreport}\\

Gemma-3-4B-it 
& Mistral-7B-Instruct 
& Granite-3.3-8B-Instruct 
& R1-Distill-Qwen-14B \\

\cite{gemmateam2025gemma3technicalreport}& \cite{jiang2023mistral7b}& & \cite{deepseekai2025deepseekr1incentivizingreasoningcapability} \\

Nemotron-Mini-4B 
& OLMo-3-7B-Instruct 
& LLaMA-3.1-8B-Instruct 
& \\

&\cite{olmo2025olmo3} & \cite{weerawardhena2025llama31foundationaisecurityllm8binstructtechnicalreport}& \\

R1-Distill-Qwen-1.5B 
& Qwen2.5-7B-Instruct 
& Ministral-8B-Instruct
& \\

\cite{deepseekai2025deepseekr1incentivizingreasoningcapability} & \cite{qwen2025qwen25technicalreport} &\cite{liu2026ministral3} & \\

Phi-4-Mini-Reasoning (4B) 
& R1-Distill-Qwen-7B 
& Mistral-NeMo-8B-Instruct 
& \\

\cite{xu2025phi4minireasoningexploringlimitssmall} &\cite{deepseekai2025deepseekr1incentivizingreasoningcapability} & & \\

& 
& RNJ-1-Instruct (8B) 
& \\

& & \cite{rnj1_instruct}& \\

&
& Qalb-1.0-8B-Instruct
& \\

& & \cite{hassan2026qalblargeststateofthearturdu}& \\

\bottomrule
\end{tabularx}

\caption{Categorization of evaluated models by parameter scale}
\label{tab:model-size-categories}
\end{table}

\section{Extended Results}
\label{app:extended_results}




\begin{figure}[h]
    \centering
    \includegraphics[width=0.8\columnwidth]{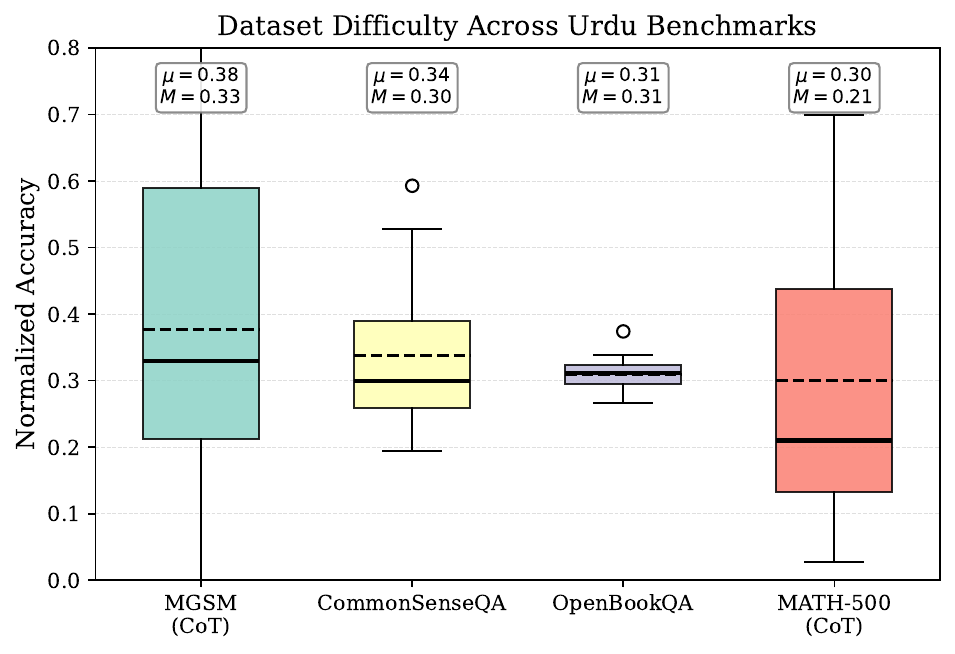}
    \caption{Distribution of model performance across four Urdu benchmarks. Box plots 
    revealing 
    MATH-500 and MGSM as the most challenging tasks with lower median scores and 
    wider performance variance compared to CommonSenseQA and OpenBookQA.}
    \label{fig:datasets_box_plot}
\end{figure}

\begin{table*}[h!]
\centering
\hspace*{-1.5cm}
\small
\renewcommand{\arraystretch}{1.4}
\caption{MATH-500 Urdu Performance by Difficulty Level}
\renewcommand{\arraystretch}{2.0}
\setlength{\tabcolsep}{8pt}
\begin{tabular}{l|ccccc|c}
\hline
\textbf{Model} & \textbf{L1} & \textbf{L2} & \textbf{L3} & \textbf{L4} & \textbf{L5} & \textbf{Avg} \\

\hline \multicolumn{7}{c}{\textit{Top-performing LLMs in Math500-Urdu}} \\ \hline

Gemma-3-4B-it & 65.1 & 58.9 & 56.2 & 39.8 & 22.4 & 44.2\\ 

Phi-4-mini-reasoning (4B) &  76.7 & 67.8 & 66.7 & 69.5 & 56.72 & 65.8\\ 

Qwen2.5-7B-Instruct & 76.7 & 56.7 & 46.7 & 32.8 & 23.1 & 41.2\\

Falcon-h1-7b-instruct & 76.7 & 61.1 & 56.2 & 46.9 & 30.6 & 49.6\\ 

DeepSeek-R1-Distill-Qwen-7B &  65.1 & 57.8 & 60.9 & 56.3 & 51.5 & 57.0\\ 

Gemma2-9B-it & 74.4 & 56.7 & 44.8 & 29.7 & 12.7 & 37.0\\ 

Gemma-3-12B-it & 86.1 & 83.3 & 80.0 & 69.5 & 48.5 & 70.0\\ 

DeepSeek-R1-Distill-Qwen-14B & 79.1 & 70.0 & 76.2 & 65.6 & 63.4 & 69.2\\ 

\hline
\end{tabular}

\label{tab:math500-levels}
\footnotesize\begin{flushleft}
~~~~~~~~Lx correspond to increasing MATH-500 difficulty levels.\\
~~~~~~~~All results are zero-shot, CoT based prompts.\\
\end{flushleft}

\end{table*}

\begin{figure*}[t!]
\centering
\begin{tikzpicture}
\begin{axis}[
    width=0.95\textwidth,
    height=0.6\textwidth,
    xlabel={Model Parameters (Billions)},
    ylabel={Average Accuracy (\%)},
    xmin=0, xmax=15,
    ymin=15, ymax=65,
    grid=major,
    grid style={dashed, gray!30},
    legend style={
        at={(0.98,0.02)},
        anchor=south east,
        fill=white,
        fill opacity=0.9,
        draw=black!50,
        rounded corners=2pt,
        font=\small,
        cells={anchor=west},
        line width=0.8pt,
    },
    tick label style={font=\footnotesize},
    label style={font=\normalsize},
    every axis plot/.append style={thick},
    xlabel style={font=\normalsize},
    ylabel style={font=\normalsize},
]

\addplot[
    only marks,
    mark=*,
    mark size=3.5pt,
    color=blue!70!black,
    fill=blue!40,
    ] coordinates {
    (1, 17.0)   
    (3, 29.7)   
    (4, 44.9)   
    (4, 24.5)   
    (7, 53.5)   
    (7, 25.6)   
    (7, 17.2)   
    (7, 24.7)   
    (7, 40.2)   
    (8, 31.4)   
    (8, 29.2)   
    (8, 25.9)   
    (8, 42.9)   
    (8, 34.6)   
    (8, 21.4)   
    (8, 27.6)   
    (9, 42.2)   
    (12, 59.4)  
    (8, 23.6) 
};

\addplot[
    only marks,
    mark=triangle*,
    mark size=4pt,
    color=red!70!black,
    fill=red!40,
    ] coordinates {
    (1.5, 19.6)  
    (4, 37.7)    
    (7, 38.3)    
    (14, 44.9)   
};

\node[font=\scriptsize\bfseries, above right, fill=green!10, draw=green!50, rounded corners=1pt, inner sep=1.5pt] 
    at (axis cs:12,59.4) {Gemma-3-12B (59.4\%)};

\node[font=\scriptsize\bfseries, above, fill=blue!10, draw=blue!50, rounded corners=1pt, inner sep=1.5pt] 
    at (axis cs:7,53.5) {Falcon-h1-7B (53.5\%)};

\node[font=\scriptsize\bfseries, below right, fill=blue!10, draw=blue!50, rounded corners=1pt, inner sep=1.5pt] 
    at (axis cs:4,44.9) {Gemma-3-4B (44.9\%)};

\node[font=\scriptsize, above left, inner sep=1.5pt] 
    at (axis cs:14,44.9) {R1-Qwen-14B};

\node[font=\scriptsize, below left, inner sep=1.5pt] 
    at (axis cs:7,38.3) {R1-Qwen-7B};

\node[font=\scriptsize, above right, inner sep=1.5pt] 
    at (axis cs:4,37.7) {Phi-4-mini};

\node[font=\scriptsize, right] at (axis cs:3,29.7) {Llama-3.2-3B};
\node[font=\scriptsize, right] at (axis cs:1,17.0) {MobileLLM-Pro};
\node[font=\scriptsize, right] at (axis cs:1.5,19.6) {R1-Qwen-1.5B};

\node[font=\scriptsize, left] at (axis cs:7,40.2) {Qwen2.5-7B};
\node[font=\scriptsize, below left] at (axis cs:7,25.6) {MiMo-7B};
\node[font=\scriptsize, left] at (axis cs:7,17.2) {Mistral-7B};
\node[font=\scriptsize, above left] at (axis cs:7,24.7) {Olmo-3-7B};

\node[font=\scriptsize, right] at (axis cs:8,23.6) {Qalb-1.0-8B};

\node[font=\scriptsize, above right] at (axis cs:8,42.9) {Llama-3.1-8B};
\node[font=\scriptsize, right] at (axis cs:8,34.6) {Ministral-8B};
\node[font=\scriptsize, right] at (axis cs:8,31.4) {Alif-1.0-8B};
\node[font=\scriptsize, left] at (axis cs:8,29.2) {Aya-Expanse};
\node[font=\scriptsize, right] at (axis cs:8,25.9) {Granite-3.3};
\node[font=\scriptsize, left] at (axis cs:8,21.4) {Mistral-NeMo};
\node[font=\scriptsize, right] at (axis cs:8,27.6) {Rnj-1};

\node[font=\scriptsize, below] at (axis cs:4,24.5) {Nemotron-4B};

\node[font=\scriptsize, below right] at (axis cs:9,42.2) {Gemma2-9B};

\addplot[
    domain=1:12,
    samples=100,
    color=blue!50,
    dashed,
    line width=1pt,
    forget plot,
] {15 + 3.5*x};

\legend{Non-Reasoning LLMs, Reasoning-Oriented LLMs}

\end{axis}
\end{tikzpicture}
\caption{Performance scaling of LLMs on Urdu benchmark datasets. The plot shows average normalized accuracy across MGSM (CoT), MATH-500 (CoT), CommonsenseQA (Direct), and OpenBookQA (Direct) as a function of model parameters. Non-reasoning models (blue circles) demonstrate stronger scaling behavior compared to reasoning-oriented models (red triangles). Gemma-3-12B (green box) achieves the highest performance at 59.4\%, while reasoning-focused models show competitive performance. Blue boxe shows top performing models in 4B and 7B model parameters The dashed line indicates the approximate scaling trend for non-reasoning models.}
\label{fig:model_scaling_2}
\end{figure*}
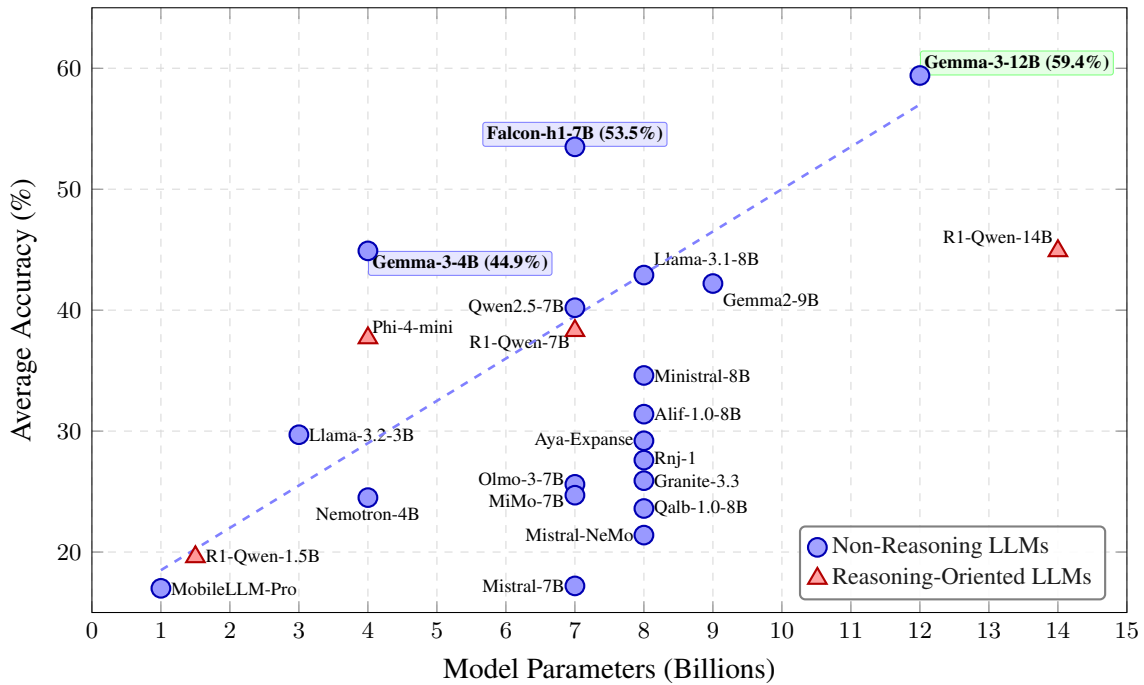

\clearpage        

\end{document}